\documentclass[10pt,twocolumn,letterpaper]{article}

\usepackage{cvpr}              

\usepackage[accsupp]{axessibility}  

\usepackage{amsmath}
\usepackage{amssymb}
\usepackage{booktabs}
\usepackage[table]{xcolor}    
\usepackage{multirow}   
\usepackage{graphicx}   
\usepackage{subcaption}
\usepackage[page, title]{appendix}
\usepackage{pifont}
\usepackage{makecell} 
\usepackage{array}      
\usepackage{mathtools} 
\usepackage{algorithm}
\usepackage{algorithmic}

\usepackage[dvipsnames, svgnames, x11names]{xcolor} 
\usepackage{graphicx} 
\usepackage{bm}
\definecolor{sourceidblue}{HTML}{0000FF}       
\definecolor{sourceoodred}{HTML}{FF0000}       
\definecolor{shiftidlightblue}{HTML}{00FFFF}    
\definecolor{shiftoodpink}{HTML}{FF00FF}      
\definecolor{compensatedidgreen}{HTML}{32CD32} 
\definecolor{compensatedoodyellow}{HTML}{FFD700} 

\definecolor{cvprblue}{rgb}{0.21,0.49,0.74}
\usepackage[pagebackref,breaklinks,colorlinks,allcolors=cvprblue]{hyperref}

\def\paperID{*****} 
\def\confName{CVPR}
\def\confYear{2026}

\title{Back to Source:  Open-Set Continual Test-Time Adaptation \\ via Domain Compensation}

\author{
Yingkai Yang \quad Chaoqi Chen\thanks{Corresponding author.} \quad Hui Huang\\
College of Computer Science and Software Engineering, Shenzhen University\\
{\tt\small \{ekyleyang, cqchen1994, hhzhiyan\}@gmail.com}
}

\begin{document}
\maketitle
\begin{abstract}
Test-Time Adaptation (TTA) aims to mitigate distributional shifts between training and test domains. 
However, existing TTA methods fall short in a realistic scenario where models face both continually changing domains and simultaneous emergence of unknown semantic classes --- a challenging setting we term Open-set Continual Test-Time Adaptation (OCTTA). The coupling of domain and semantic shifts often collapses the feature space, severely degrading both classification and out-of-distribution detection. 
To tackle this, we propose DOmain COmpensation (\textbf{DOCO}), an effective framework that robustly performs domain adaptation and OOD detection in a synergistic closed loop. DOCO first performs dynamic, adaptation-conditioned sample splitting to separate likely ID from OOD samples. Using only the ID samples, it learns a \textbf{domain compensation prompt} by aligning feature statistics with the source domain, guided by a structural regularizer that prevents semantic distortion. This learned prompt is then propagated to the OOD samples \textbf{within} the same batch, isolating their semantic novelty for reliable detection. Extensive experiments on multiple benchmarks show that DOCO outperforms prior continual and open-set TTA methods, establishing a new state-of-the-art for OCTTA. Code is released at \url{https://github.com/ekyle0522/DOCO}.

\begin{figure}[t]
\centering
\includegraphics[width=1.0\columnwidth]{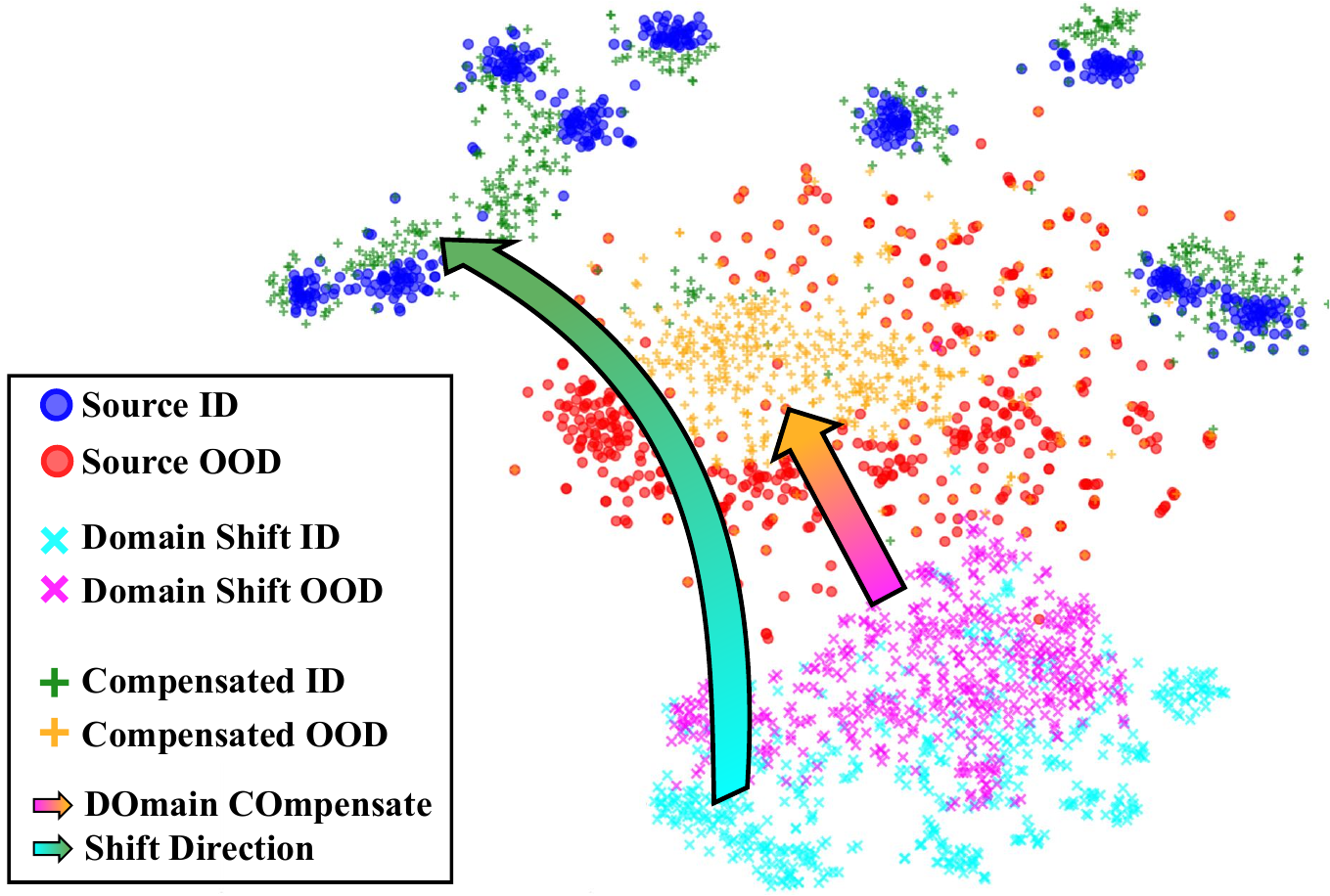} 
\caption{
    \textbf{Illustration of ``Back to Source''}.
    This t-SNE plot shows that while source ID (\textcolor{sourceidblue}{\textbullet}) and OOD (\textcolor{sourceoodred}{\textbullet})  features are well-separated, a severe domain shift collapses their feature space, mixing ID and OOD features into an inseparable cluster (\textcolor{shiftidlightblue}{$\times$} and \textcolor{shiftoodpink}{$\times$}).
    By applying \textbf{DO}main \textbf{CO}mpensation method, the features of compensated ID (\textcolor{compensatedidgreen}{$+$}) and OOD (\textcolor{compensatedoodyellow}{$+$})  samples  are successfully realigned with the original source structure.
    This disentangles domain and semantic shifts, making corrupted data clearly separable again.
}
\label{fig1:tsne_cover_page}
\end{figure}

\end{abstract}    
\section{Introduction}

\begin{figure}[t]
\centering
\includegraphics[width=0.88\columnwidth]{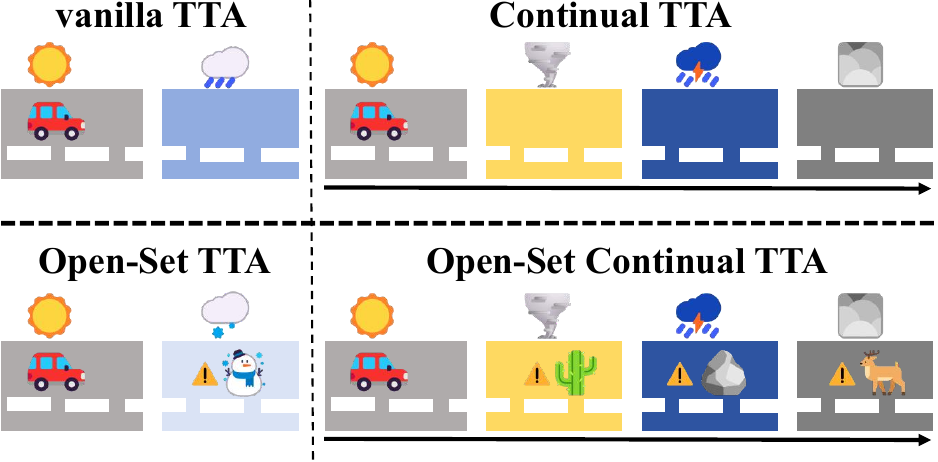} 
\caption{Illustration of different TTA settings.}
\label{fig2: TTAsettings}
\end{figure}

Deploying pre-trained models in the real world invariably confronts domain shift, where the test distribution differs from the source and degrades performance.
To mitigate this, test-time adaptation (TTA) adapts a source-trained model on the fly using only unlabeled target data at inference.
Beyond early single target domain setting \cite{wang2021tent}, recent studies highlight two realistic axes: continual TTA, which copes with nonstationary streams \cite{wang2022continual,niu2022efficient,zhang2025dpcore,liu2024vida}, and open-set TTA, where unknown classes co-occur with shifted known ones \cite{lee2023towards,gao2024unified,yu2024stamp,zhang2025come}. We focus on their intersection and formulate the Open-set Continual Test-Time Adaptation (OCTTA) scenario that stresses both stability and unknown-awareness, as depicted in \cref{fig2: TTAsettings}.
For instance, a visual perception system deployed in the wild must adapt not only to changing environments, such as from a clear highway scene to a foggy forest scene (domain shift), but also to unexpected objects appearing in those scenes, such as a deer on the road (semantic shift). Crucially, these novel objects are subject to the \textbf{same} domain shifts as objects from the known classes, creating a coupled challenge. Robustly handling these combined shifts is therefore important for reliable real-world visual systems.

The OCTTA setting poses a tripartite challenge for existing methods. First, the continuous stream of domain shifts exacerbates catastrophic forgetting, eroding knowledge of the source domain as the model adapts to new ones \cite{wang2022continual}. Second, the mixture of in-distribution (ID) and out-of-distribution (OOD)\footnote{In this paper, ID refers to data within the source semantic space, OOD refers to data outside it.} samples corrupts the batch statistics required by normalization-based methods and can misguide the optimization in entropy minimization-based approaches \cite{gao2024unified}. Third, and most critically, is the \textbf{antagonistic coupling} of domain and semantic shifts (\cref{fig1:tsne_cover_page} and Appendix \ref{sec：sup_easyhard}). A severe domain shift can collapse the feature space, ``squashing'' the embeddings of both known and unknown classes into a poorly separable region. This collapse blurs class boundaries, simultaneously crippling the model's ability to classify known data and to detect novelties.

To address these challenges, we propose \textbf{DO}main \textbf{CO}mpensation, a simple and intuitive framework for OCTTA setting. DOCO integrates domain adaptation and OOD detection into a unified, cyclical process. The cycle begins with \textbf{Back-to-Source Prompt Learning}, where a lightweight domain compensation prompt is updated using likely ID samples from the current online batch. This update, performed via one-step backpropagation, aligns target feature statistics with those of the source domain to neutralize the domain shift without distorting semantics. This learned prompt, which now encodes the current domain information, is then immediately applied to the likely OOD samples in the same batch through \textbf{Intra-Batch Prompt Propagation}. By compensating for their domain shift, this step isolates their semantic novelty, making them more distinguishable from known classes. The loop is sustained and improved by \textbf{Adaptation-Conditioned Sample Splitting}: prompt-adapted features enable more accurate ID/OOD separation for the current batch, yielding a purer ID set for prompt learning and better capture of the current domain mode. Improved splitting thus creates a virtuous cycle, enhancing prompts and, in turn, future adaptation and detection. This closed-loop design makes the system resilient to continuous and unforeseen environmental changes.

Our main contributions are as follows:
\begin{itemize}
    \item We formally introduce Open-set Continual Test-Time Adaptation (OCTTA), a pragmatic setting that reflects real-world complexities. We show that prior TTA methods suffer performance degradation caused by coupled domain and semantic shifts in the feature space.
    \item We propose DOCO, a visual prompt learning-based framework that effectively mitigates the negative coupling of shifts through domain compensation and establishes a positive feedback loop via a dynamic sample-splitting mechanism.
    \item We conduct extensive experiments on multiple OOD datasets under the OCTTA setting. The results show that DOCO achieves state-of-the-art performance, especially surpassing the next-best UniEnt by $4.7\%$ on ImageNet-C, validating its effectiveness and robustness.
\end{itemize}

\section{Related Work}

\begin{figure*}[t]
\centering
\includegraphics[width=0.95\textwidth]{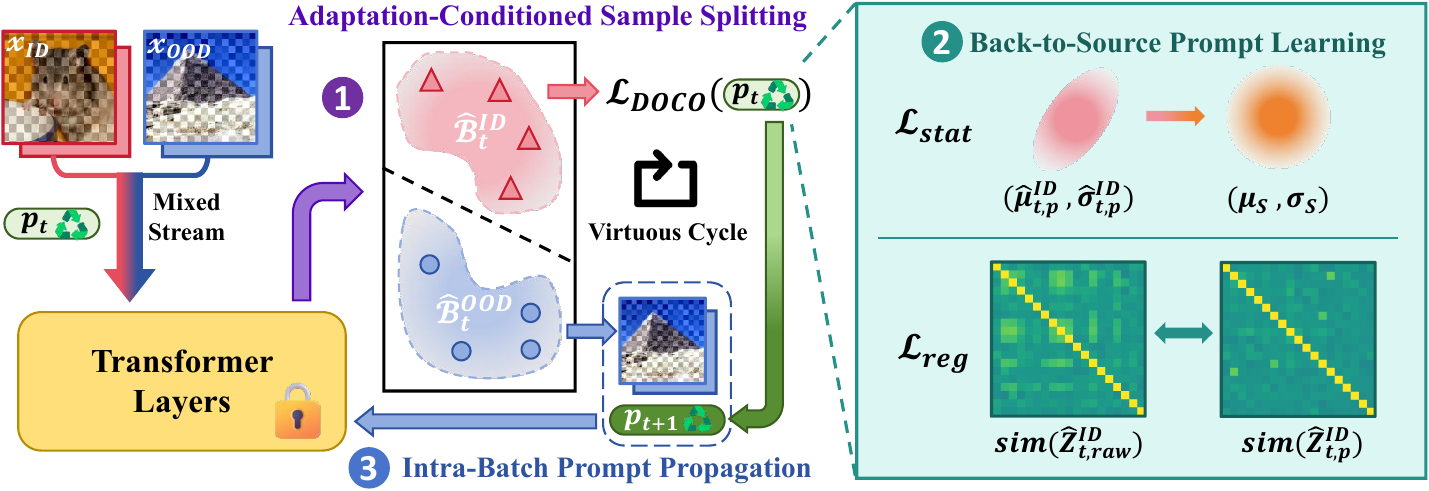}

\caption{
Overview of DOCO framework.
\ding{182}~The incoming batch is processed with the prompt \(p_{t}\) to reduce domain shift, and features $z_{i,p}$ are scored against source class prototypes \(W_h = \{w_c\}_{c \in \mathcal{Y}^S}\) to split the batch into  $\hat{\mathcal{B}}_t^{\text{ID}}$ and $\hat{\mathcal{B}}^\text{OOD}_{t}$.
\ding{183}~A new prompt \(p_{t+1}\) is optimized on the ID subset by minimizing \(\mathcal{L}_{\text{DOCO}}\) in one-step backpropagation.
\ding{184}~The newly optimized prompt \(p_{t+1}\) is immediately propagated to OOD data within the same batch, neutralizing domain shift for accurate OOD detection.
}
\label{fig3:main_pipeline}
\end{figure*}

\subsection{Test-time Adaptation}
TTA \cite{dong2025aeo,wang2023dynamically,lee2024entropy,iwasawa2021test,huselective,li2023robustness} adapts a source-trained model online without labeled target data. We summarize methods along the two axes introduced above. For continual TTA, CoTTA \cite{wang2022continual} stabilizes long-horizon updates via weight/augmentation-averaged targets with stochastic neuron restoration. EATA \cite{niu2022efficient} filters unreliable or redundant samples and regularizes important weights to curb risky updates and forgetting. SAR \cite{niu2023towards} replaces brittle BN with batch-agnostic norms and employs sharpness-aware reliable entropy to avoid collapse under wild shifts. ViDA \cite{liu2024vida} introduces lightweight adapters to decouple domain-shared from domain-specific factors. DPCore \cite{zhang2025dpcore} retains domain knowledge through a dynamic prompt coreset aimed at recurring or short-lived domains. For open-set TTA, wisdom-of-crowds filtering \cite{lee2023towards} suppresses losses whose confidence decreases after adaptation. UniEnt \cite{gao2024unified} jointly minimizes entropy on pseudo-csID and maximizes it on pseudo-csOOD with marginal-entropy regularization. STAMP \cite{yu2024stamp} leverages a stable, class-balanced memory with self-weighted entropy. COME \cite{zhang2025come} regularizes confidence via conservative entropy minimization to enhance open-world stability.

\subsection{Out-of-Distribution Detection}
OOD detection \cite{tao2023nonparametric,du2022vos,wang2023outofdistribution,zhu2023diversified} separates ID from unknown samples without domain shift. Two widely used lines are post-hoc scoring and test-time detection. For post-hoc scoring, MSP \cite{hendrycks17baseline} thresholds predicted confidence, and energy-based scoring \cite{liu2020energy} replaces confidence with log-sum-exp energy and often surpasses MSP and MaxLogit \cite{Hendrycks2022ScalingOD}. For test-time detection, RTL \cite{fan2024test} learns a lightweight linear map from features to OOD scores directly at inference. AUTO \cite{yang2023auto} performs online optimization with an in–out-aware filter, an ID memory, and a consistency loss. CODA \cite{chen2023coda} compacts source embeddings with virtual unknowns and disambiguates known from unknown through prototype-guided updates. Recent dictionary-based designs such as OODD  \cite{yang2025oodd} maintain a dynamic OOD feature dictionary and calibrate scores via feature–dictionary similarity during testing.

\section{Preliminaries}
In this section, we provide a brief overview of Vision Transformers (ViTs) and the Visual Prompt Tuning (VPT) paradigm, and then formally define the OCTTA setting.

\vspace{-5pt}
\paragraph{ViTs \& VPT.}
We use the ViT-base architecture \cite{dosovitskiy2020vit} as our backbone.
A model $f$ with parameters $\theta$ is decomposed as a feature extractor $\phi:\mathcal{X}\to\mathcal{Z}$ with parameters $\theta_\phi$ and a linear classifier $h:\mathcal{Z}\to\mathbb{R}^{C}$ with parameters $\theta_h$, i.e., $f_\theta = h \circ \phi$. Concretely, we write $h(z)=W_h z + b_h,$
where $W_h=[w_1,\ldots,w_C]^\top$ and $h_c(z)$ denotes the logit for class $c$. Given an image, we denote by $z\in\mathcal{Z}$ the $[\texttt{CLS}]$ representation after the last transformer block, and obtain predictions via $\hat{y}=\arg\max_{c\in[C]} h_c(z)$.
For efficient test-time adaptation, we adopt Visual Prompt Tuning (VPT) \cite{jia2022visual}. We augment the input sequence with \(L\) learnable prompt tokens \(p=\{\texttt{[Prompt]}_i\}_{i=0}^{L-1}\), inserted after \([\texttt{CLS}]\) and before the patch tokens, yielding the first-layer input \(\{[\texttt{CLS}],\, p,\, \text{patch}_1,\ldots,\text{patch}_k\}\). During adaptation, we freeze the original model parameters \(\theta\)
and update only the prompts \(p\).
This parameter-efficient design greatly reduces the number of trainable parameters and mitigates catastrophic forgetting, while still allowing the model to swiftly adjust to shifting test distributions.

\vspace{-5pt}
\paragraph{OCTTA Problem Formulation.} Given a model \( f_\theta \) pre-trained on a source domain \( \mathcal{D}^S = (\mathcal{X}^S, \mathcal{Y}^S) \), the goal is to adapt \( f_\theta \) to a sequence of unlabeled target domains \( \{\mathcal{D}^{T_i}\}_{i=1}^M \). Each target domain \( \mathcal{D}^{T_i} \) possesses its label space \( \mathcal{Y}^{T_i} \), and the source label space is a proper subset of the target, i.e., \( \mathcal{Y}^S \subset \mathcal{Y}^{T_i} \). The model receives an online stream of test batches \( \{\mathcal{B}_t\}_{t=1}^\infty \). For any domain \( \mathcal{D}^{T_i} \) , we model the data distribution \( P_{\text{test}}^{T_i} \) with the Huber contamination model \cite{huber1992robust} to represent coupled domain and semantic shifts:
\begin{equation}
    P_{\text{test}}^{T_i} = (1 - \kappa) P_{\text{ID-C}}^{T_i} + \kappa P_{\text{OOD-C}}^{T_i}.
\label{eq:mixed_distribution}
\end{equation}
In this mixture, \( \kappa \in [0,1) \) is a preset ratio for OOD data. While $\kappa$ controls the overall OOD proportion, each mini-batch is randomly sampled, resulting in a stochastic ID-to-OOD ratio per batch to better simulate real-world data streams. \( P_{\text{ID-C}}^{T_i} \) is the distribution of ID samples \( (x, y) \) with known class label where \( y \in \mathcal{Y}^S \), while \( P_{\text{OOD-C}}^{T_i} \) is the distribution of OOD samples with novel classes where \( y \in \mathcal{Y}^{T_i} \setminus \mathcal{Y}^S \). The `\texttt{-C}' suffix signifies that both ID and OOD data are affected by the same domain corruption specific to \( \mathcal{D}^{T_i} \). This adaptation process unfolds online, where at each time step \( t \), the model's parameters \( \theta_t \) are updated to \( \theta_{t+1} \) using only the current batch \( \mathcal{B}_t \) and a single backpropagation step. The objective is to correctly classify future samples from known classes \( \mathcal{Y}^S \) while detecting those from novel classes \( \mathcal{Y}^{T_i} \setminus \mathcal{Y}^S \).

\section{Methodology}

\paragraph{Overview.} We introduce DOCO, a novel framework for the OCTTA setting, with an overview provided in  \cref{fig3:main_pipeline}. DOCO consists of three major components and systematically addresses a series of core challenges: (1) How the prompt can adapt to the target domain without semantic overfitting (\cref{sec:back_learning})?  (2) How to reuse the learned domain knowledge to bootstrap the model's inference capabilities (\cref{sec:OODinference})? and (3) How to effectively separate ID and OOD samples under severe domain shifts (\cref{sec:method:splitting})? These components operate within a synergistic closed-loop, where more accurate splitting enables better prompt learning, which in turn improves inference and guides future adaptation steps. Algorithm is detailed in Appendix \ref{sec:sup_alg}.

\subsection{Back-to-Source Prompt Learning}
\label{sec:back_learning}
Upon receiving a batch $\mathcal{B}_t$ at time $t$, we first isolate a subset of likely ID samples, denoted as \( \hat{\mathcal{B}}_t^{\mathrm{ID}} \subset \mathcal{B}_t \), using a dynamic splitting mechanism detailed in \cref{sec:method:splitting}. For this subset, we optimize the current prompt \( p_{t} \) and obtain an updated prompt \( p_{t+1} \). The purpose of the prompt is to align the feature distribution of $\hat{\mathcal{B}}_t^{\mathrm{ID}}$ with that of the source domain $\mathcal{D}^S$, thereby counteracting the domain shift while preserving the intrinsic semantic structure of the samples.

\vspace{-10pt}
\paragraph{Statistical Alignment.} 
A domain shift from \( \mathcal{D}^S \) to \( \mathcal{D}^{T_i} \) will inevitably cause a shift in the statistics of the feature space \cite{NIPS2006_b1b0432c}. 
We argue that encouraging the ID feature statistics to move back toward the source statistics enables the prompt to approximate a compensation for the domain shift in latent space \cite{NEURIPS2024_d3602fc9}.
To this end, we pre-cache the mean \( \mu_S \) and standard deviation \( \sigma_S \) of the source features calculated from a small set of unlabeled samples in an offline manner. For the current ID batch \( \hat{\mathcal{B}}_t^{\mathrm{ID}} \), we extract the features using the prompt \( p_{t} \), yielding \( \hat{Z}_{t,p}^{\mathrm{ID}} = \{ \phi(x; p_{t}) \}_{x \in \hat{\mathcal{B}}_t^{\mathrm{ID}}} \). We then compute the batch statistics, mean \( \hat{\mu}_{t,p}^{\mathrm{ID}} \) and standard deviation \( \hat{\sigma}_{t,p}^{\mathrm{ID}} \). The statistical alignment loss is defined as the L2 distance between the batch and source statistics:
\begin{equation}
\label{eq:Lstat}
    \mathcal{L}_{\mathrm{stat}}(p_{t}) = \| \hat{\mu}_{t,p}^{\mathrm{ID}} - \mu_S \|_2 + \| \hat{\sigma}_{t,p}^{\mathrm{ID}} - \sigma_S \|_2.
\end{equation}

\vspace{-10pt}
\paragraph{Structural Preservation.} Relying solely on $\mathcal{L}_{\mathrm{stat}}$ is insufficient, as batch statistics reflect both domain shift and batch-specific semantics. For instance, if a batch contains only ``dogs" and ``cats", forcing its feature statistics to match the entire source statistics with far more classes could compel the prompt to distort the feature structure, causing it to overfit to the batch's narrow semantics rather than learning a general domain compensation.

To address this, we introduce a regularization term to preserve the relative feature geometry. Specifically, we enforce that the pairwise similarity structure within the selected ID subset remains consistent before and after applying the prompt. 
Let $\hat{\mathcal{B}}_t^{\mathrm{ID}}=\{x_i\}_{i=1}^{n}$, where $n=|\hat{\mathcal{B}}_t^{\mathrm{ID}}|$. For each $x_i$, let
$z_i^{\mathrm{raw}}=\phi(x_i)$ and $z_i^{p_t}=\phi(x_i;p_t)$
denote its raw and prompted feature representations, respectively. Accordingly,
$\hat{Z}_{t,\mathrm{raw}}^{\mathrm{ID}}=\{z_i^{\mathrm{raw}}\}_{i=1}^{n}$ and
$\hat{Z}_{t,p}^{\mathrm{ID}}=\{z_i^{p_t}\}_{i=1}^{n}$.
We use cosine similarity
\begin{equation}
C(a,b)
=
\frac{a \cdot b}{\|a\|_2 \, \|b\|_2},
\label{eq:cosine}
\end{equation}
which is evaluated on both raw features $(z_i^{\mathrm{raw}}, z_j^{\mathrm{raw}})$ and prompted features $(z_i^{p_t}, z_j^{p_t})$. The structural preservation loss is then defined as the Frobenius norm of the difference between the two pairwise similarity matrices:
\begin{equation}
\mathcal{L}_{\mathrm{reg}}(p_t)
=
\sqrt{
\sum_{i=1}^{n}\sum_{j=1}^{n}
\left(
C(z_i^{p_t}, z_j^{p_t})
-
C(z_i^{\mathrm{raw}}, z_j^{\mathrm{raw}})
\right)^2
},
\label{eq:reg_expanded}
\end{equation}
or equivalently,
\begin{equation}
\mathcal{L}_{\mathrm{reg}}(p_t)
=
\left\|
\mathrm{sim}(\hat{Z}_{t,p}^{\mathrm{ID}})
-
\mathrm{sim}(\hat{Z}_{t,\mathrm{raw}}^{\mathrm{ID}})
\right\|_F,
\label{eq:Lreg_compact}
\end{equation}
where for a feature set $Z=\{z_i\}_{i=1}^{n}$, $\mathrm{sim}(Z)\in\mathbb{R}^{n\times n}$ denotes its pairwise cosine-similarity matrix, whose $(i,j)$-th entry is $C(z_i,z_j)$. By penalizing disruptions to the relative feature geometry, this regularizer encourages the prompt to compensate for domain shift without overfitting to the narrow semantics of the current batch. The final objective for optimizing the current prompt \(p_t\) is
\begin{equation}
\mathcal{L}_{\mathrm{DOCO}}(p_t)
=
\mathcal{L}_{\mathrm{stat}}(p_t)
+
\beta \mathcal{L}_{\mathrm{reg}}(p_t),
\label{eq:L_DOCO}
\end{equation}
where \( \beta \) is a regularization coefficient. The prompt parameters are then optimized on the current ID subset to obtain an updated prompt \(p_{t+1}\).

\begin{figure*}[t]
\centering 
\begin{subfigure}[b]{0.25\textwidth} 
    \centering
    \includegraphics[width=\linewidth]{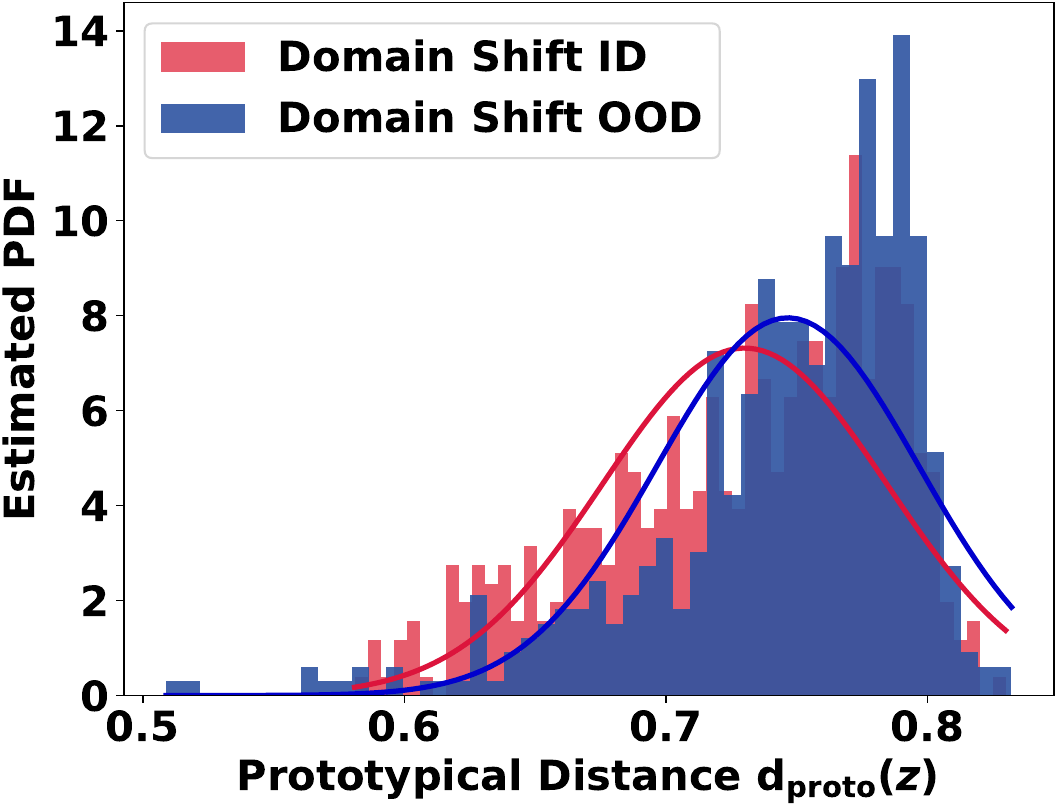} 
    \caption{ w/o prompt.}
    \label{fig:split1} 
\end{subfigure}
\begin{subfigure}[b]{0.243\textwidth} 
    \centering
    \includegraphics[width=\linewidth]{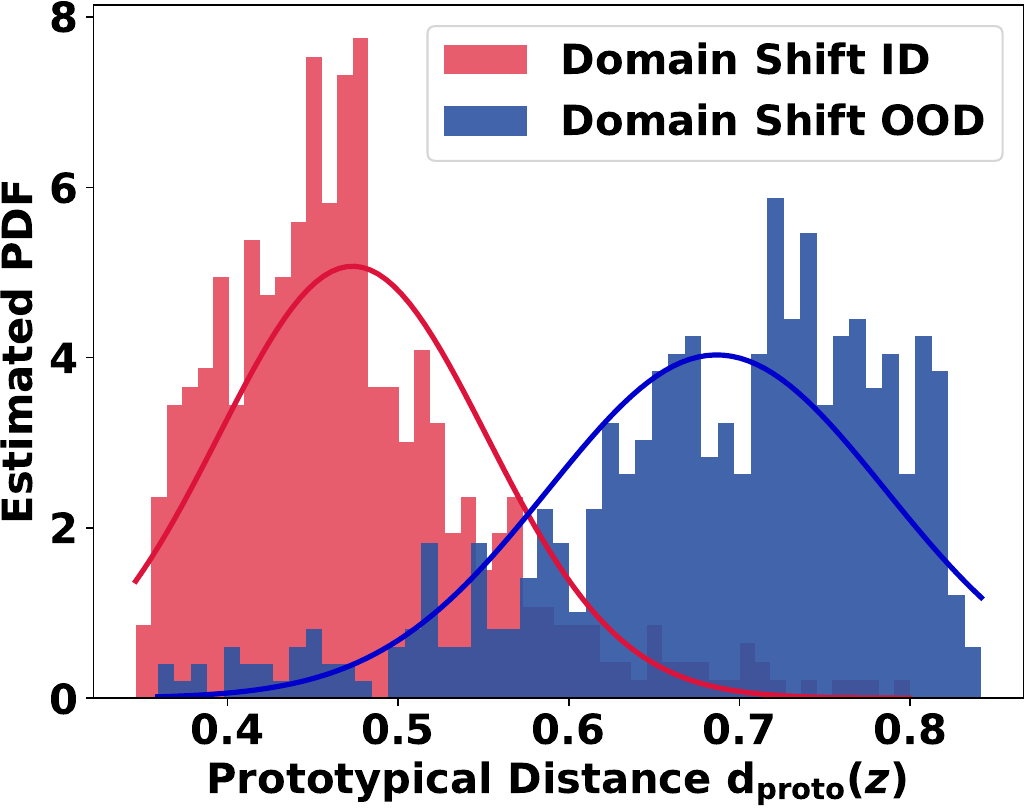} 
    \caption{ w/ prompt.}
    \label{fig:split2} 
\end{subfigure}
\begin{subfigure}[b]{0.49\textwidth} 
    \centering
    \includegraphics[width=\linewidth]{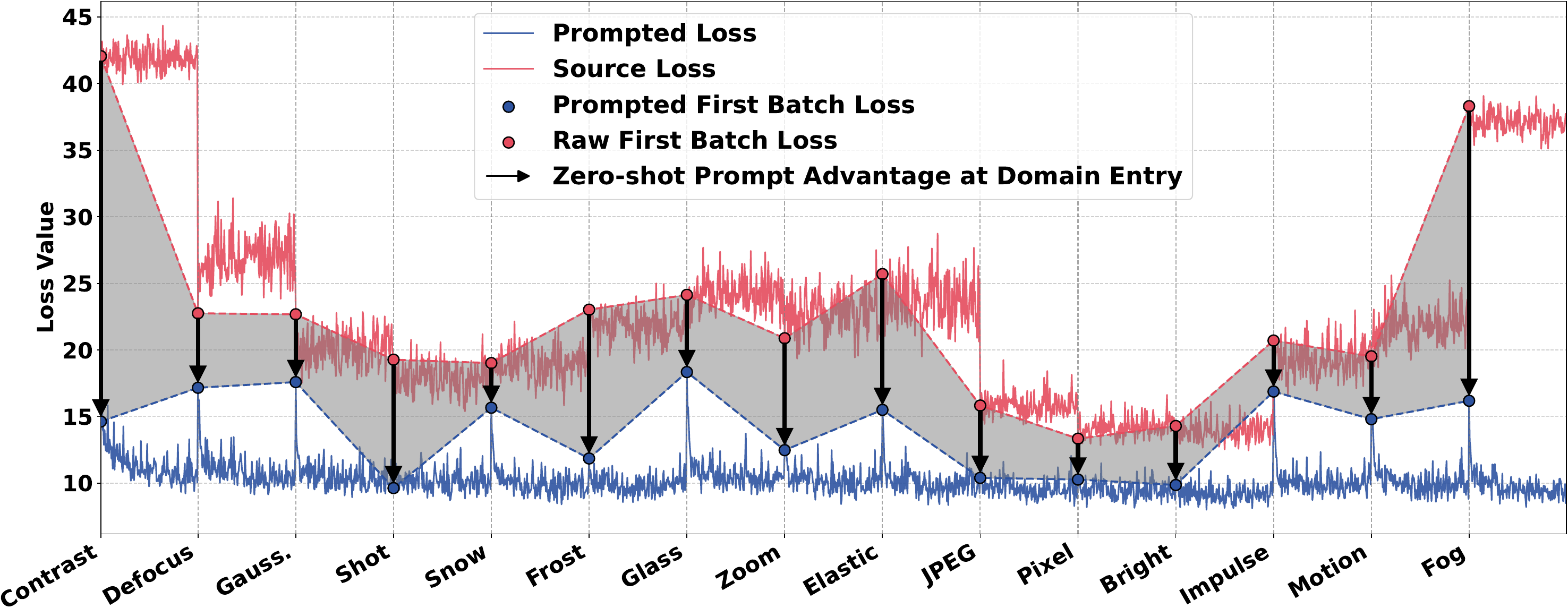} 
    \caption{$\mathcal{L}_{\mathrm{stat}}$ across 15 domains w/ and w/o prompt.}
    \label{fig:lossgap} 
\end{subfigure}

\caption{(a-b) DOCO facilitates the separation of ID and OOD by disentangling the bimodal distribution of the prototypical distance $d_{\mathrm{proto}}(z)$. (c) For each new domain, the statistical loss $\mathcal{L}_{\mathrm{stat}}(p_t)$ with DOCO is consistently smaller than that of the source model on the first batch, indicating that the learned prompt generalizes well to unseen domains.}
\label{fig5:bimodal_and_gap} 
\end{figure*}

\subsection{Intra-Batch Prompt Propagation}
\label{sec:OODinference}
\noindent\textbf{Direct Intra-Batch Knowledge Reuse.}
Within the current batch, samples in $\hat{\mathcal{B}}_t^{\mathrm{ID}}$ are inferred using the prompt $p_t$. We then immediately reuse the batch-$t$ domain knowledge learned from these likely ID samples --- instantiated as the updated prompt $p_{t+1}$ --- by applying it only to likely OOD samples in $\hat{\mathcal{B}}_t^{\mathrm{OOD}}$ within the same batch.
Since all samples in $\mathcal{B}_t$ share the same batch-wise domain component $\delta_t$, the prompt $p_{t+1}$ compensates the domain factor of OOD features in a consistent manner. The final prediction for likely OOD samples is produced by the frozen linear classifier head $h$ on compensated features:
\begin{equation}
\hat y \;=\; \arg \max_{c\in \mathcal{Y}^S} \, h_c\!\big(\phi(x; p_{t+1})\big),
\qquad x \in \hat{\mathcal{B}}_t^{\mathrm{OOD}}.
\label{eq:ood_prediction}
\end{equation}

\noindent\textbf{Motivation.}
Recent OOD generalization work \cite{miao2025dics} suggests that domain-invariant semantics can be exposed by removing domain-specific components from representations. 
As an intuition, we write the batch-$t$ representation as
$\phi(x)\approx s(x)+\delta_t$, where $s(x)$ encodes class semantics and $\delta_t$ is a batch-wise domain factor.
We learn $p_{t+1}$ from only likely ID samples in $\mathcal{B}_t$ and then \emph{immediately} propagate it to the likely OOD subset in the same batch, yielding
\begin{equation}
    \phi(x; p_{t+1}) \;\approx\; \phi(x) - \delta_t \;\approx\; s(x).
\end{equation}
This propagation is non-trivial: it uses ID-only updates to estimate and neutralize the \emph{same} batch factor for all samples sharing $\delta_t$, which (i) corrects mis-split IDs by pulling them back toward source-aligned neighborhoods, (ii) makes true OODs more novel relative to the compensated source geometry, and (iii) avoids leaking pseudo-label noise by not back-propagating through likely OOD samples, thus stabilizing the decision boundary for the whole batch. Comparison between training-time explicit separation \cite{miao2025dics} and our test-time in-process correction is provided in Appendix \ref{sec:sup_three-pathways-compact}.

\subsection{Adaptation-Conditioned Sample Splitting}
\label{sec:method:splitting}

\paragraph{Domain overshadowing and prompt generalization.}
The effectiveness of our framework relies on splitting each batch \(\mathcal{B}_t\) into likely ID and likely OOD subsets, \(\hat{\mathcal{B}}_t^{\mathrm{ID}}\) and \(\hat{\mathcal{B}}_t^{\mathrm{OOD}}\).
This separation prevents OOD contamination during prompt learning, leading to a cleaner estimation of the batch-wise domain patterns.
However, severe domain shifts can overshadow semantic differences  (\cref{fig:split1}), causing substantial overlap between the ID and OOD distributions of the prototypical distance \(d_{\mathrm{proto}}\) (defined in Eq.~\ref{eq:proto-distance}).
Therefore, for each batch we first compute compensated features
\(Z_{t,p}=\{\phi(x; p_t)\}_{x\in\mathcal{B}_t}\) using the prompt $p_t$.
Even under new domains, the structure-preserving prompt exhibits strong cross-domain generalization (\cref{fig:lossgap} and Appendix \ref{sec:sup_prompt-generalization}), enabling DOCO to address the ``Continual'' aspect of O\textbf{C}TTA effectively.
This “back-to-source” effect restores a clearer bimodality (\cref{fig:split2}), enabling reliable partitioning.

\vspace{-10pt}

\paragraph{Prototypical Distance Splitting.}
Given the prompted features \(Z_{t,p}\) and the frozen classifier weights \(\{w_c\}_{c\in\mathcal{Y}^S}\) serving as source prototypical proxies~\cite{gong2025out}, for each \(z\in Z_{t,p}\) we define the \emph{prototypical distance}
\vspace{-5pt}
\begin{equation}
\label{eq:proto-distance}
d_{\mathrm{proto}}(z) \;=\; 1 \;-\; \max_{c\in\mathcal{Y}^S} \, C(z, w_c).
\end{equation}
Here, \(C(\cdot,\cdot)\) denotes cosine similarity defined in Eq.~\ref{eq:cosine}.
Let \(\mathcal{S}_t \coloneqq \{\, d_{\mathrm{proto}}(z)\mid z\in Z_{t,p}\,\}\) denote the collection of prototypical distances induced by this batch. We run \(K\)-Means with $K=2$ over these scalar scores:
\vspace{-3pt}
\begin{equation}
\label{eq:kmeans-compact}
\begin{aligned}
(\widehat{\mathcal{K}}_0,\widehat{\mathcal{K}}_1)
&= \arg\min_{\mathcal{K}_0,\mathcal{K}_1}\; 
\sum_{i\in\{0,1\}}\;\sum_{d\in \mathcal{K}_i}\|d-\mu_i\|^2, \\
\text{s.t.}\quad 
& \mathcal{K}_0\cup \mathcal{K}_1=\mathcal{S}_t,\;\; 
\mathcal{K}_0\cap \mathcal{K}_1=\varnothing .
\end{aligned}
\end{equation}
where $\mu_i = \frac{1}{|\mathcal{K}_i|}\sum_{d\in\mathcal{K}_i} d$ denotes the centroid of cluster $\mathcal{K}_i$.
Since a smaller \(d_{\mathrm{proto}}\) indicates being closer to the source prototypes, 
we assign the cluster with the smaller centroid to ID. 
Denoting that index by \(i_{\mathrm{ID}}\in\{0,1\}\) and \(i_{\mathrm{OOD}}=1-i_{\mathrm{ID}}\), the split is
\vspace{-3pt}
\begin{equation}
\label{eq:split-compact}
\begin{aligned}
\hat{\mathcal{B}}_t^{\mathrm{ID}}  &= \big\{\,x\in\mathcal{B}_t \;\big|\; d_{\mathrm{proto}}(\phi(x;p_t)) \in \widehat{\mathcal{K}}_{i_{\mathrm{ID}}}\big\},\\
\hat{\mathcal{B}}_t^{\mathrm{OOD}} &= \big\{\,x\in\mathcal{B}_t \;\big|\; d_{\mathrm{proto}}(\phi(x;p_t)) \in \widehat{\mathcal{K}}_{i_{\mathrm{OOD}}}\big\}.
\end{aligned}
\end{equation}

\section{Experiments}

\begin{table*}[t]
\centering
\small
\setlength{\tabcolsep}{3.8pt}
\caption{Results ($\%$) for ImageNet-to-ImageNet-C (severity = 5, $\kappa = 0.5$) in OCTTA setting across six covariate-shifted OOD datasets. All the results are \textbf{averaged over 15} corruptions. \textbf{Bold} and \underbar{underline} are used to indicate the first and second best performance, respectively. }
\renewcommand{\arraystretch}{0.8}
\begin{tabular}{c cc cc cc cc cc cc ccc}
\toprule
\multirow{2}{*}{\textbf{Method}} &
\multicolumn{2}{c}{\textbf{Places.--C}} &
\multicolumn{2}{c}{\textbf{Texture--C}} &
\multicolumn{2}{c}{\textbf{iNatur.--C}} &
\multicolumn{2}{c}{\textbf{SUN--C}} &
\multicolumn{2}{c}{\textbf{SSB-H.--C}} &
\multicolumn{2}{c}{\textbf{NINCO--C}} &
\multicolumn{3}{c}{\textbf{Avg.}} \\
\cmidrule(lr){2-3} \cmidrule(lr){4-5} \cmidrule(lr){6-7} \cmidrule(lr){8-9} \cmidrule(lr){10-11} \cmidrule(lr){12-13} \cmidrule(lr){14-16}
& ACC & AUC & ACC & AUC & ACC & AUC & ACC & AUC & ACC & AUC & ACC & AUC & ACC & AUC & H-score \\
\midrule
Source & 49.8 & 66.8 & 49.8 & 70.5 & 49.8 & 78.7 & 49.8 & 71.4 & 49.8 & 56.3 & 49.8 & 64.5 & 49.8 & 68.0 & 56.4 \\
Tent (ICLR'21) \cite{wang2021tent} & 12.3 & 49.3 & 3.8 & 44.6 & 12.1 & 50.1 & 13.3 & 38.9 & 55.3 & 62.5 & 37.5 & 59.8 & 22.4 & 50.9 & 23.8 \\
CoTTA (CVPR'22) \cite{wang2022continual} & 49.9 & 63.0 & 48.8 & 65.5 & 49.4 & 75.3 & 49.6 & 65.8 & 49.9 & 55.5 & 49.4 & 62.0 & 49.5 & 64.5 & 54.8 \\
EATA (ICML'22) \cite{niu2022efficient} & 54.9 & 65.4 & 50.5 & 72.1 & 51.9 & 72.9 & 52.1 & 70.8 & 56.3 & 58.3 & 51.8 & 64.3 & 52.9 & 67.3 & 57.8 \\
SAR (ICLR'23) \cite{niu2023towards} & 51.2 & 59.6 & 45.0 & 60.0 & 46.0 & 64.3 & 49.5 & 57.9 & 56.8 & 60.5 & 53.7 & 66.8 & 50.4 & 61.5 & 54.3 \\
OSTTA (ICCV'23) \cite{lee2023towards} & 57.5 & 60.2 & 56.5 & 55.5 & 50.9 & 66.3 & 55.6 & 59.4 & 58.8 & 62.5 & 57.8 & 67.2 & 56.2 & 61.9 & 58.5 \\
ViDA (ICLR'24) \cite{liu2024vida} & 53.0 & 59.8 & 52.1 & 36.8 & 53.1 & 35.9 & 51.5 & 30.4 & 55.2 & 61.9 & 53.2 & 62.9 & 53.0 & 47.9 & 48.4 \\
UniEnt (CVPR'24) \cite{gao2024unified} & \underbar{58.6} & 74.1 & 55.5 & \underbar{79.7} & 56.8 & \underbar{89.6} & 58.1 & \underbar{84.6} & 59.1 & 62.1 & \underbar{58.6} & 72.2 & 57.8 & \underbar{77.0} & \underbar{65.4} \\
STAMP (ECCV'24) \cite{yu2024stamp} & 51.9 & 72.4 & 52.0 & 75.1 & 52.1 & 85.6 & 51.9 & 77.4 & 52.0 & 60.9 & 51.9 & 71.4 & 51.9 & 73.8 & 60.2 \\
E-COME (ICLR'25) \cite{zhang2025come} & 57.7 & 76.0 & \underbar{57.8} & 78.6 & \underbar{58.9} & 85.4 & \underbar{59.3} & 82.7 & 59.2 & 58.9 & 56.6 & 71.3 & \underbar{58.3} & 75.5 & 65.2 \\
S-COME (ICLR'25) \cite{zhang2025come} & 31.2 & 58.5 & 29.3 & 67.1 & 22.9 & 55.6 & 54.6 & 76.4 & 54.3 & 55.6 & 53.2 & 65.1 & 40.9 & 63.0 & 45.5 \\
DPCore (ICML'25) \cite{zhang2025dpcore} & 56.4 & \underbar{76.7} & 54.3 & 78.0 & 46.7 & 82.8 & 52.6 & 82.0 & \underbar{60.2} & \underbar{62.6} & 54.3 & \underbar{75.3} & 54.1 & 76.2 & 62.6 \\
\rowcolor{blue!10}
\textbf{DOCO (Ours)} & \textbf{61.8} & \textbf{80.6} & \textbf{61.0} & \textbf{84.6} & \textbf{61.4} & \textbf{95.7} & \textbf{61.5} & \textbf{92.2} & \textbf{61.9} & \textbf{65.9} & \textbf{61.5} & \textbf{77.4} & \textbf{61.5} & \textbf{82.7} & \textbf{70.1} \\
\bottomrule
\end{tabular}
\label{tab1:imgNetC}
\end{table*}

\subsection{Experimental Setup}

\paragraph{Datasets.} For ID component, we first use the standard benchmark ImageNet-C \cite{hendrycks2019robustness}.
This dataset contains 15 common corruption types, and we use the highest severity level 5. To further assess performance, we then adopt a newly-released LAION-C benchmark \cite{li2025laionc}, which produces six challenging, synthetic distortions.
Given their significant difficulty, we use a moderate severity level of 3 for these corruptions. For the OOD component, we follow common practices but apply the same corruption to standard OOD benchmarks. Specifically, the OOD datasets are corrupted versions of Places365 \cite{zhou2017places}, Textures \cite{cimpoi2014describing}, NINCO \cite{bitterwolf2023ninco}, iNaturalist \cite{van2018inaturalist}, SSB-Hard \cite{vaze2022openset}, and SUN \cite{xiao2010sun}. 

\vspace{-10pt}
\paragraph{Baselines.}  
We compare our method against strong continual and open-set TTA baselines including Tent \cite{wang2021tent}, CoTTA \cite{wang2022continual}, SAR \cite{niu2023towards}, EATA \cite{niu2022efficient}, ViDA \cite{liu2024vida}, DPCore \cite{zhang2025dpcore}, STAMP \cite{yu2024stamp}, OSTTA \cite{lee2023towards}, UniEnt \cite{gao2024unified}, and COME \cite{zhang2025come}.

\vspace{-10pt}
\paragraph{Evaluation Protocol.}

We use three metrics for a comprehensive evaluation. First, we report ID classification accuracy \textbf{(ACC)} to measure domain generalization. Second, we use the area under the ROC curve \textbf{(AUC)} to assess the model's threshold-free outlier detection capability. Finally, to jointly evaluate both aspects, we report the \textbf{H-score}, which is the harmonic mean of ACC and AUC.

\begin{figure}[t]
\centering
\newlength{\outgrow}
\setlength{\outgrow}{0.02\columnwidth} 

\begin{subfigure}[b]{0.5\columnwidth}
  \makebox[\linewidth][l]{%
    \hspace*{-\outgrow}%
    \includegraphics[width=\dimexpr\linewidth+\outgrow\relax]{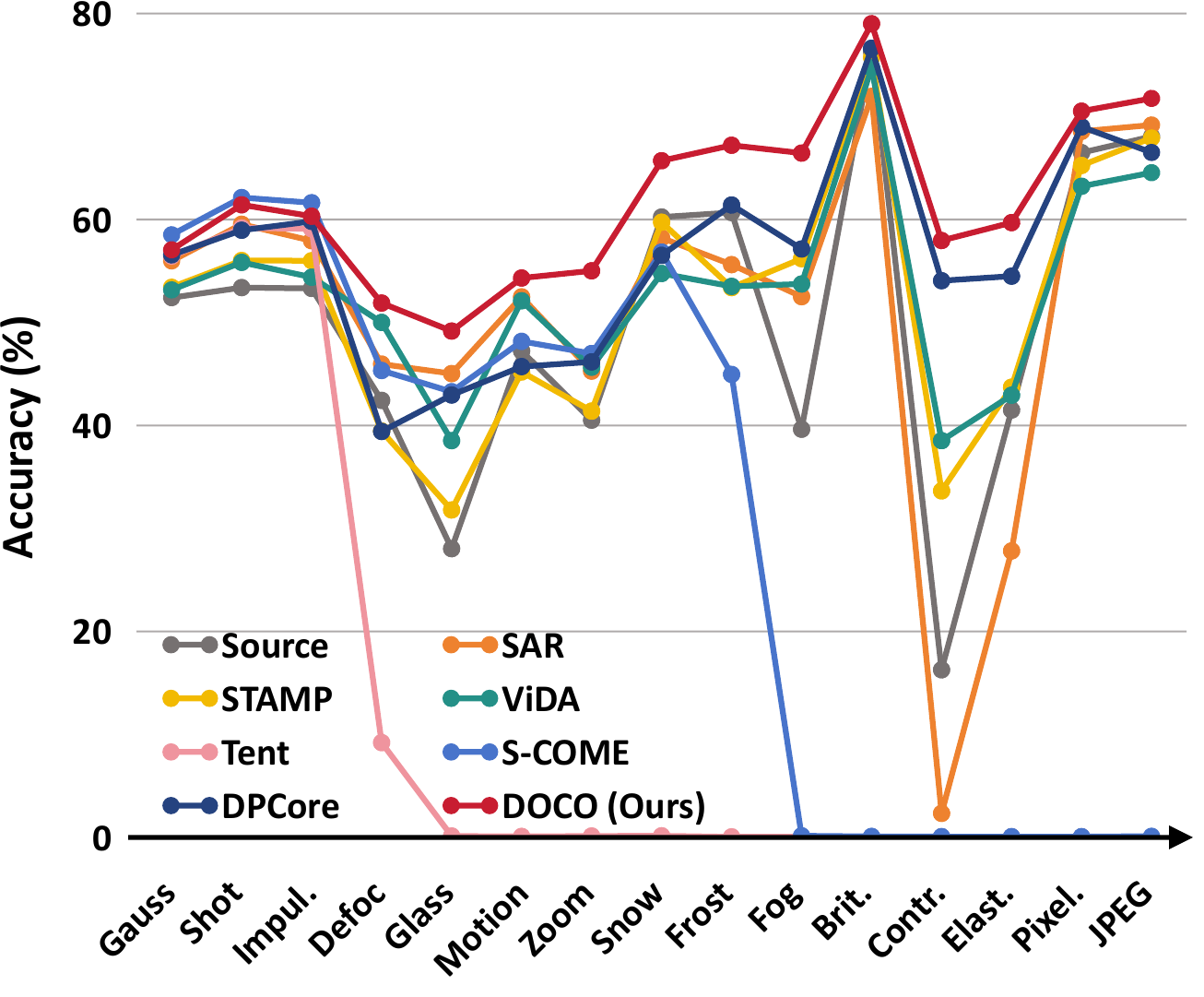}}
  \caption{Cross domain Acc}
\label{fig6a:line_graph}
\end{subfigure}%
\begin{subfigure}[b]{0.5\columnwidth}
  \makebox[\linewidth][r]{%
    \includegraphics[width=\dimexpr\linewidth+\outgrow\relax]{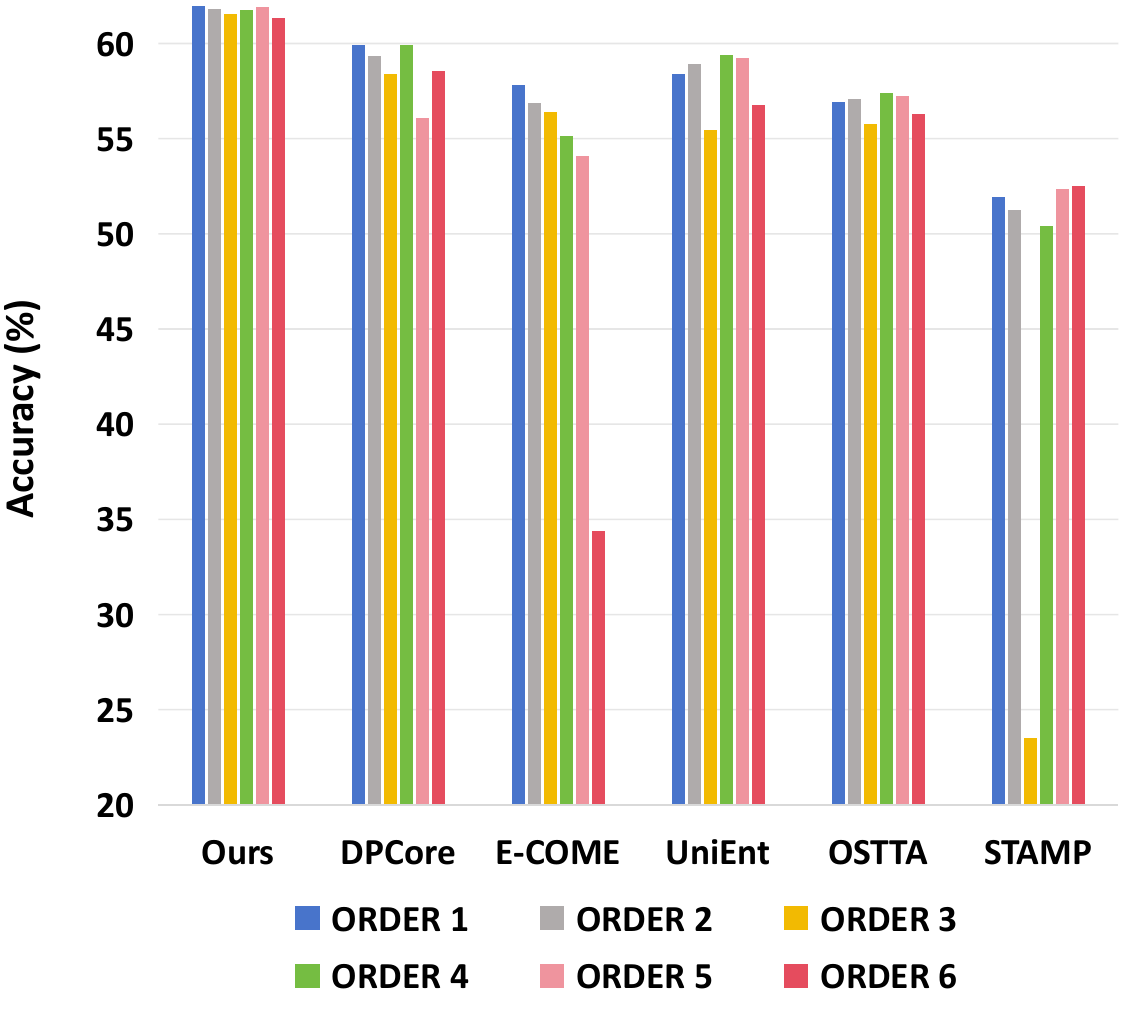}%
    \hspace*{-\outgrow}}%
  \caption{Different domain order}
    \label{fig6b:bar_graph}
\end{subfigure}
\caption{Further analysis of the OCTTA setting on the ImageNet-C benchmark. (a) Per-domain accuracy ($\%$) of different models under continuous domain shifts. (b) Accuracy ($\%$) of models tested with different domain orders. }
\label{fig:lineandbar}
\end{figure}

\vspace{-10pt}
\paragraph{Implementation Details.}
Experiments use a ViT-B/16 pretrained on ImageNet-1K, with weights from \texttt{timm}. For evaluating OOD detection, we use the energy score \cite{liu2020energy} to compute AUC. The test stream follows Eq.~\ref{eq:mixed_distribution} with OOD ratio $\kappa=0.5$, and we standardize the batch size to 64 for fair comparison.
To simulate a real-world deployment and ensure a fair blind test, hyperparameters for all methods are tuned only on the first dataset-domain combination. These settings are then frozen and applied to all subsequent, unseen datasets to rigorously evaluate generalization robustness \cite{yu2024stamp}.
Prior to online adaptation, we pre-compute source feature statistics from 300 samples, following EATA \cite{niu2022efficient}. The prompts conduct a one-time self-supervised update for 50 iterations to refine the initial state. During TTA, we update $L=8$ learnable prompts for each incoming batch using AdamW (learning rate $1\text{e-}1$) with structural preservation weight $\beta=0.5$.
In contrast to methods like CoTTA which update all model parameters, DOCO exclusively fine-tunes the prompts, while most other baselines update only the affine parameters of LayerNorm layers. For the plug-in frameworks UniEnt and COME, we evaluate their \textbf{strongest} open-set and continual configurations: EATA with UniEnt+ (full version, denoted as \texttt{UniEnt} for short), EATA with COME (\texttt{E-COME}) and SAR with COME (\texttt{S-COME}). More details are available in Appendix \ref{sec:sup_baseline}.

\begin{figure}[t]
\centering

\setlength{\outgrow}{0.02\columnwidth} 

\begin{subfigure}[b]{0.5\columnwidth}
  \makebox[\linewidth][l]{%
    \hspace*{-\outgrow}
    \includegraphics[width=\dimexpr\linewidth+\outgrow\relax]{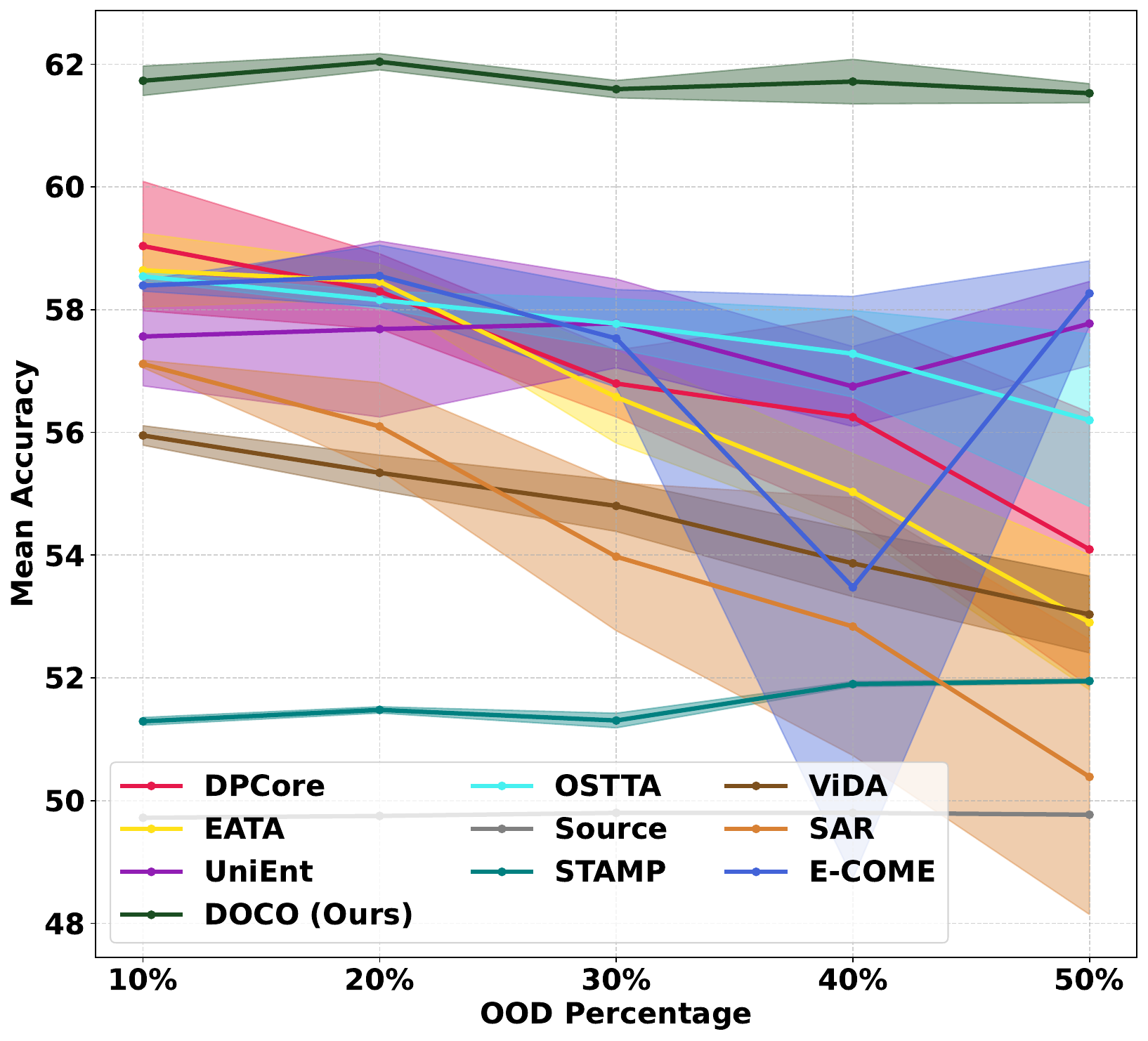}}
  \caption{Accuracy (\%)}
  \label{fig:std_acc}
\end{subfigure}%
\begin{subfigure}[b]{0.5\columnwidth}
  \makebox[\linewidth][r]{%
    \includegraphics[width=\dimexpr\linewidth+\outgrow\relax]{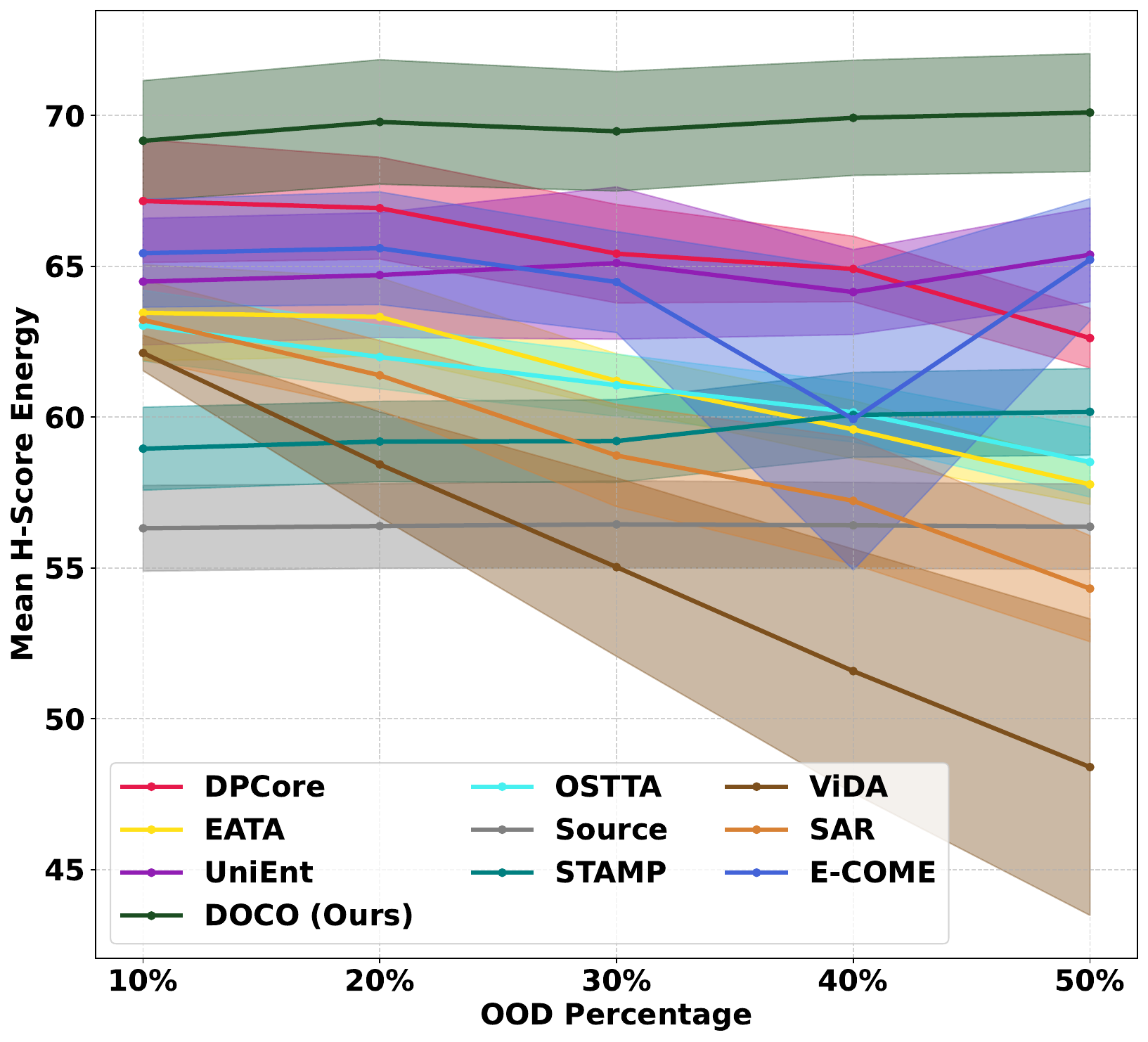}%
    \hspace*{-\outgrow}}
  \caption{H-score (\%)}
  \label{fig:std_hs}
\end{subfigure}

\caption{Performance under different OOD ratios.}
\label{fig:stdline}
\end{figure}

\subsection{Main Results}

\begin{table*}[t]
\centering
\small
\setlength{\tabcolsep}{3.8pt}
\caption{Results ($\%$) for LAION-C benchmark (severity = 3, $\kappa = 0.5$) in the OCTTA setting across six domain shift OOD datasets. All results are averaged over 6 constantly switching domains. `\texttt{-L}' stands for applying LAION-C corruption to OOD dataset. }
\renewcommand{\arraystretch}{0.8}
\begin{tabular}{c cc cc cc cc cc cc ccc}
\toprule
\multirow{2}{*}{\textbf{Method}} &
\multicolumn{2}{c}{\textbf{Places.--L}} &
\multicolumn{2}{c}{\textbf{Texture--L}} &
\multicolumn{2}{c}{\textbf{iNatur.--L}} &
\multicolumn{2}{c}{\textbf{SUN--L}} &
\multicolumn{2}{c}{\textbf{SSB-H.--L}} &
\multicolumn{2}{c}{\textbf{NINCO--L}} &
\multicolumn{3}{c}{\textbf{Avg.}} \\
\cmidrule(lr){2-3} \cmidrule(lr){4-5} \cmidrule(lr){6-7} \cmidrule(lr){8-9} \cmidrule(lr){10-11} \cmidrule(lr){12-13} \cmidrule(lr){14-16}
& ACC & AUC & ACC & AUC & ACC & AUC & ACC & AUC & ACC & AUC & ACC & AUC & ACC & AUC & H-score \\
\midrule
Source & 16.2 & 52.2 & 16.2 & 54.8 & 16.2 & 58.9 & 16.2 & 54.9 & 16.2 & 54.8 & 16.2 & 55.8 & 16.2 & 55.2 & 19.1 \\
Tent (ICLR'21) & 0.2 & 47.4 & 0.2 & 58.9 & 0.4 & 46.5 & 0.2 & 49.8 & 0.4 & 49.3 & 0.2 & 53.4 & 0.3 & 50.9 & 0.6 \\
CoTTA (CVPR'22) & 16.2 & 50.7 & 16.1 & 51.7 & 16.1 & 56.1 & 16.1 & 52.8 & 16.1 & 54.5 & 16.1 & 54.8 & 16.1 & 53.4 & 18.5 \\
EATA (ICML'22) & 23.9 & 56.0 & 23.3 & 59.9 & 20.7 & 59.6 & 22.5 & 60.3 & \underbar{25.0} & 54.4 & 23.1 & 55.1 & 23.1 & 57.5 & 27.9 \\
SAR (ICLR'23) & 7.6 & 50.6 & 9.4 & 51.8 & 4.5 & 47.9 & 4.6 & 48.4 & 13.6 & 54.4 & 5.1 & 51.5 & 7.5 & 50.8 & 9.6 \\
OSTTA (ICCV'23) & 16.3 & 49.4 & 16.7 & 48.6 & 15.6 & 47.6 & 14.5 & 48.6 & 17.1 & 54.4 & 16.9 & 53.7 & 16.2 & 50.4 & 17.5 \\
ViDA (ICLR'24) & 2.5 & 42.7 & 3.2 & 48.6 & 2.7 & 36.5 & 2.0 & 40.4 & 2.2 & 49.0 & 2.4 & 47.1 & 2.5 & 44.0 & 3.9 \\
UniEnt (CVPR'24) & \underbar{24.0} & 58.3 & \underbar{23.8} & \underbar{64.5} & \underbar{23.7} & \underbar{66.0} & \underbar{23.6} & \underbar{64.5} & 24.4 & 54.4 & 23.1 & 57.0 & \underbar{23.8} & 60.8 & 29.3 \\
STAMP (ECCV'24) & 16.4 & 53.5 & 15.6 & 54.1 & 15.3 & 56.5 & 16.0 & 53.8 & 16.2 & \underbar{55.0} & 15.7 & 55.9 & 15.9 & 54.8 & 19.5 \\
E-COME (ICLR'25) & 4.6 & 53.1 & 17.9 & 57.6 & 16.5 & 63.6 & 18.1 & 61.4 & 18.8 & 52.3 & 17.3 & 56.4 & 15.5 & 57.4 & 19.9 \\
S-COME (ICLR'25) & 0.1 & 46.4 & 0.1 & 55.4 & 0.2 & 47.6 & 0.1 & 48.8 & 0.1 & 53.3 & 0.1 & 54.2 & 0.1 & 51.0 & 0.3 \\
DPCore (ICML'25) & 23.4 & \underbar{61.6} & 21.1 & 63.3 & 20.6 & 65.5 & \textbf{25.0} & 64.2 & 24.8 & \textbf{55.6} & \underbar{23.9} & \textbf{59.8} & 23.1 & \underbar{61.7} & \underbar{30.3} \\
\rowcolor{blue!10}
\textbf{DOCO (Ours)} & \textbf{28.1} & \textbf{62.2} & \textbf{24.4} & \textbf{64.8} & \textbf{25.7} & \textbf{78.2} & 23.2 & \textbf{71.2} & \textbf{26.1} & 53.7 & \textbf{24.6} & \underbar{59.2} & \textbf{25.4} & \textbf{64.9} & \textbf{32.7} \\

\bottomrule
\end{tabular}
\label{tab2:imgNetL}
\end{table*}

\begin{table}[t]
  \centering
  \small 
  \setlength{\tabcolsep}{4pt} 
  \caption{Accuracy (\%) under the closed-set setting.}
  \label{tab:closet}
  \renewcommand{\arraystretch}{0.8}
  \begin{tabular}{ccccccc}
    \toprule
    Method & IN-C & IN-A & IN-R & IN-Sketch & IN-L & Avg. \\
    \midrule
    Source & 49.8 & 28.1 & 43.6 & 46.6 & 16.2 & 36.9 \\
    Tent \cite{wang2021tent} & 58.1 & 29.8 & 44.1 & 46.6 & 0.84 & 35.9 \\
    CoTTA \cite{wang2022continual} & 50.1 & 28.2 & 43.6 & 46.7 & 16.2 & 37.0 \\
    EATA  \cite{niu2022efficient} & 59.9 & 29.9 & 46.0 & 47.6 & 23.4 & 41.4 \\
    OSTTA  \cite{lee2023towards} & 58.2 & 29.7 & 44.0 & 46.6 & 16.9 & 39.1 \\
    ViDA  \cite{liu2024vida} & 56.5 & 29.5 & 42.8 & 45.6 & 6.0 & 36.1 \\
    UniEnt  \cite{gao2024unified} & 59.1 & 29.3 & 45.4 & 47.6 & 22.5 & 40.8 \\
    STAMP  \cite{yu2024stamp} & 51.7 & 28.8 & 43.6 & 46.7 & 16.2 & 37.4 \\
    E-COME \cite{zhang2025come} & 59.0 & \underbar{31.2} & \underbar{46.3} & \underbar{49.2} & 15.5 & 40.2 \\
    DPCore  \cite{zhang2025dpcore} & \textbf{61.7} & 29.3 & 46.3 & 49.0 & \underbar{24.5} & \underbar{42.2} \\
     
    \rowcolor{blue!10}
\textbf{DOCO (Ours)} & \underbar{60.6} & \textbf{31.7} & \textbf{47.4} & \textbf{49.5} & \textbf{26.3} & \textbf{43.1} \\
    \bottomrule
  \end{tabular}
\end{table}
\paragraph{ImageNet-C Benchmark.}
The comprehensive results in \cref{tab1:imgNetC} show that DOCO achieves state-of-the-art performance across multiple OOD datasets. On average\footnote{ACC/AUC/H-score are first computed per dataset and then averaged.}, our method sets a new SOTA with an H-score of $\bm{70.1\%}$, surpassing the next-best method UniEnt by a significant $\bm{4.7\%}$. This top-tier result is based on a leading known-class accuracy of $\bm{61.5\%}$ and an exceptional OOD detection AUC of $\bm{82.7\%}$. This demonstrates DOCO's balanced enhancement of both ID classification and OOD detection. To further analyze DOCO's robustness, we examine its performance across individual domains and different domain orderings in \cref{fig:lineandbar}. \Cref{fig6a:line_graph} tracks per-domain accuracy as the model sequentially adapts to 15 corruptions. DOCO's performance curve is consistently at or near the top, showcasing its stable adaptation. This stability is particularly pronounced on difficult domains like \textit{Contrast}, where most other methods suffer a sharp decline in accuracy. DOCO, by effectively decoupling the severe domain shift from the underlying semantic information, remains strong under such challenging conditions. Furthermore, we assess its resilience to the sequence of domains. As depicted in \cref{fig6b:bar_graph}, DOCO's Accuracy remains stable across six different random domain orderings.
This result underscores that DOCO's effectiveness does not depend on a favorable domain sequence, confirming its robustness in truly continual settings. We further investigate how the OOD percentage $\kappa$ impacts our method. As shown in \cref{fig:stdline}, we manipulate the ratio ranging from $10\%$ to $50\%$, and present the corresponding Accuracy and H-scores. The outcomes are averaged over six OOD datasets, with transparent bands indicating the standard deviation. While most baselines suffer from a significant performance drop and huge fluctuations as the ratios and datasets vary, DOCO shows the consistently superior and stable performance all the time.

\vspace{-10pt}
\paragraph{LAION-C Benchmark.} To explore the limits of adaptation under more extreme domain shifts, we evaluate our method on the recently released LAION-C benchmark \cite{li2025laionc}.
As shown in \cref{tab2:imgNetL}, the severity of these shifts causes a drastic performance degradation across all methods, highlighting the benchmark's difficulty. Even in this adversarial setting, DOCO once again demonstrates its superior robustness and establishes a new state-of-the-art. Our method achieves the highest average H-score of $\bm{32.7\%}$, leading the closest competitor by $\bm{2.4\%}$. For dataset details and low-severity experiments, please refer to Appendix \ref{sec:sup_laion} and \ref{sec:sup_laionsev1}.

\vspace{-10pt}
\paragraph{Closed-set CTTA.}  Additionally, we examine the methods under the fundamental closed-set scenario in \cref{tab:closet}. Apart from ImageNet-C and LAION-C, we introduce ImageNet-A \cite{hendrycks2021nae}, ImageNet-R \cite{hendrycks2021many} and ImageNet-Sketch \cite{wang2019learning} datasets, which provide natural adversarial, rendition and sketch domain shift. DOCO shows the best results on almost all datasets, highlighting its capability to handle a broad range of domain shifts, while also demonstrating that the design of DOCO does not harm performance in non-open-set scenarios. This outcome makes DOCO a potent competitor in the vanilla TTA and CTTA arena.

\subsection{Analysis}

\begin{table}[t]
    \small
    \caption{Ablation results for DOCO's modules.}
    \label{tab:ablation_compact_fixed_v2}
    \setlength{\tabcolsep}{3pt}
    \renewcommand{\arraystretch}{0.8}
    \begin{tabular}{l ccc cc cc}
    \toprule
    \multirow{2}{*}{\textbf{Method}} & \multicolumn{3}{c}{\textbf{Module}} & \multicolumn{2}{c}{\textbf{N (Norm-based)}} & \multicolumn{2}{c}{\textbf{P (DOCO)}} \\
    \cmidrule(r){2-4} \cmidrule(lr){5-6} \cmidrule(l){7-8}
    & \textbf{S} & \textbf{O} & \textbf{R} & \textbf{H-score}$\uparrow$ & \textbf{Gain}$\uparrow$ & \textbf{H-score}$\uparrow$ & \textbf{Gain}$\uparrow$ \\
    \midrule
    Source    & -          & -          & -          & 56.4  & -     & 56.4  & - \\
    w/o S.O.R & -          & -          & -          & 60.7  & +4.3  & 64.0  & +7.6  \\
    w/o S.O   & -          & -          & \checkmark & 62.5  & +6.1  & 67.6  & +11.2  \\
    w/o O.R   & \checkmark & -          & -          & 62.5  & +6.1  & 65.1  & +8.7  \\
    w/o O     & \checkmark & -          & \checkmark & 63.9  & +7.5 & 65.9  & +9.5  \\
    w/o R     & \checkmark & \checkmark & -          & 64.2  & +7.8  & 68.5  & +12.1  \\
    \midrule
    \rowcolor{blue!10}
\textbf{Full Model}    & \checkmark & \checkmark & \checkmark & \textbf{68.1}  & +11.7 & \textbf{70.1}  & +13.7  \\
    \bottomrule
    \end{tabular}
\end{table}%

\begin{figure*}[t]
\centering
\begin{subfigure}[b]{0.27\textwidth}
  \includegraphics[width=\linewidth]{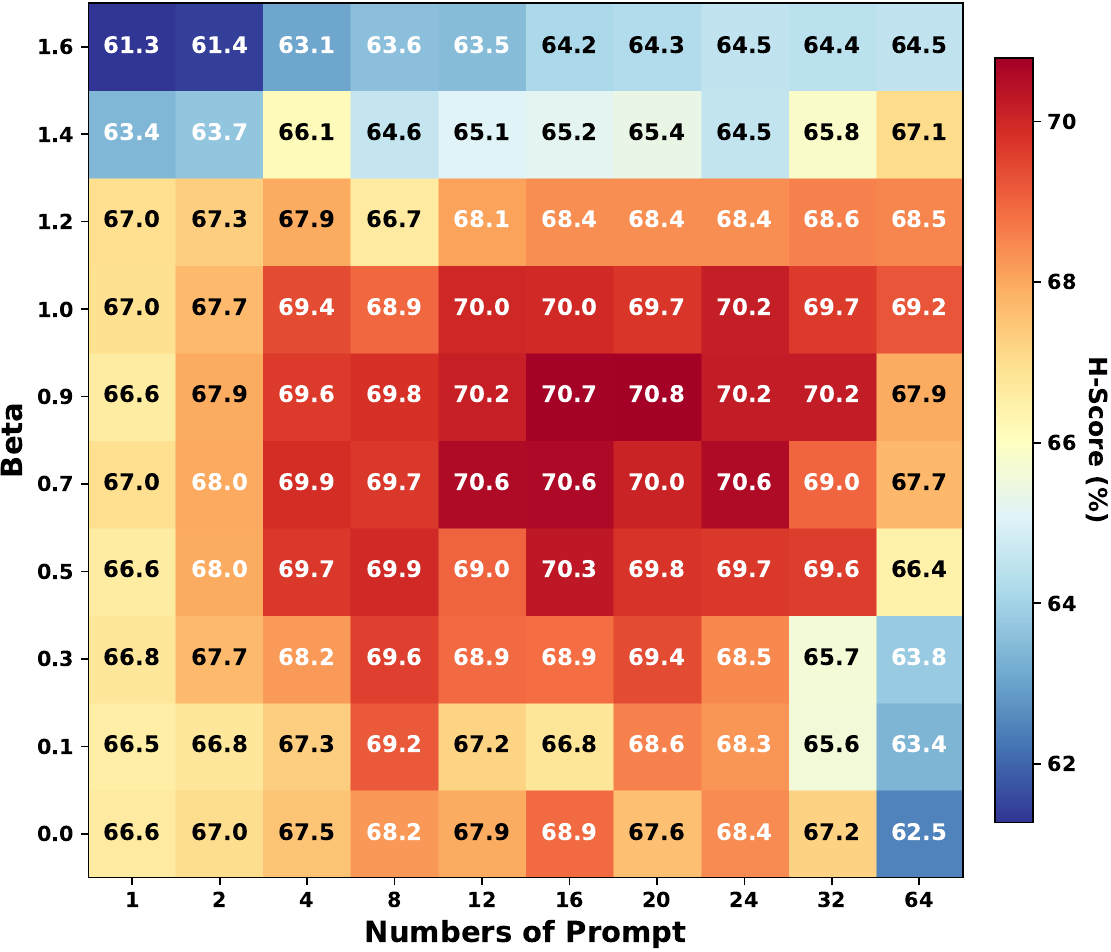}
  \caption{Sensitivity of $\beta$ vs $L$}
  \label{fig:heatmap}
\end{subfigure}
\hfill
\begin{subfigure}[b]{0.27\textwidth}
  \includegraphics[width=\linewidth]{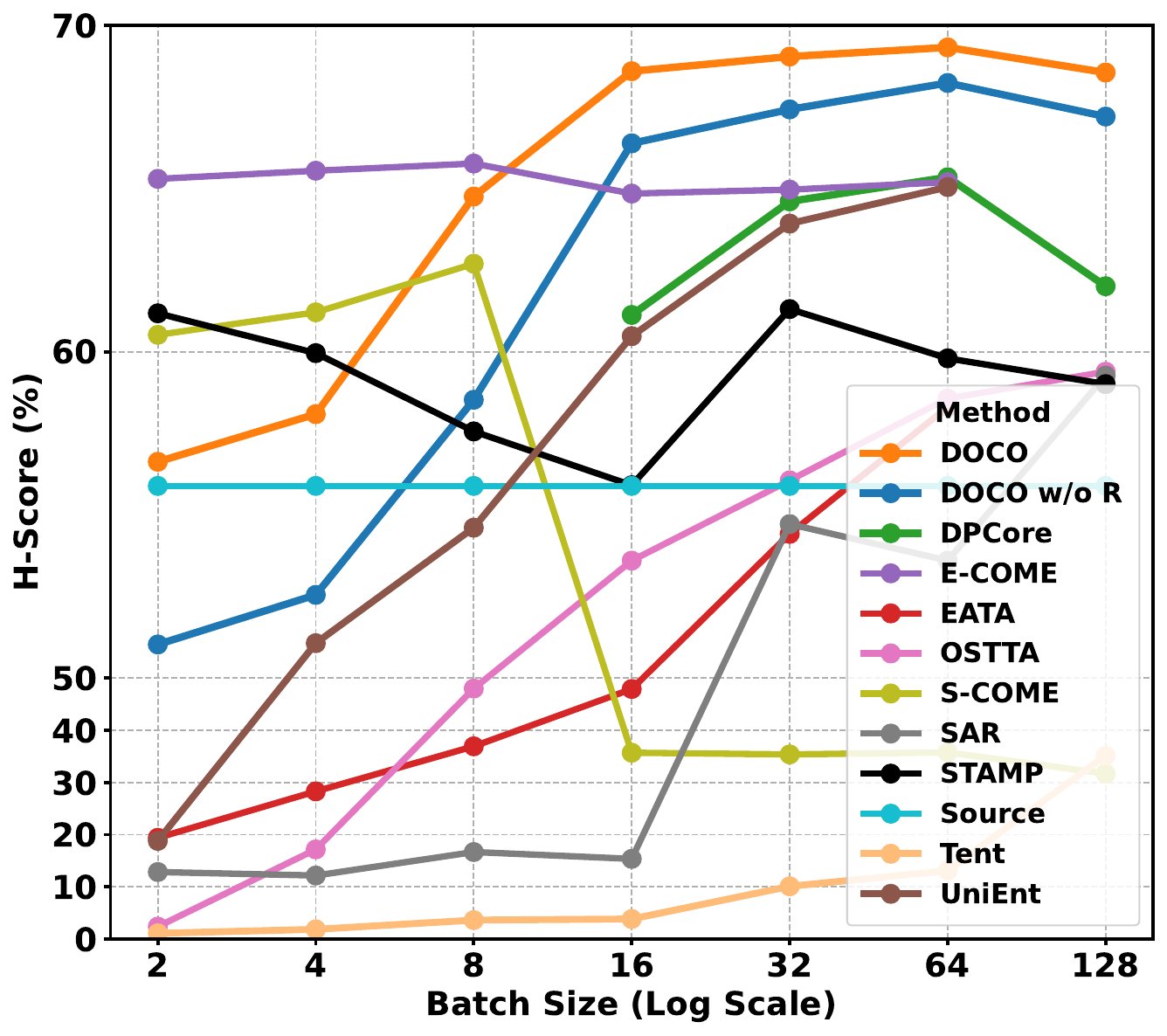}
  \caption{Batch size comparison}
  \label{fig:bsline}
\end{subfigure}
\hfill
\begin{subfigure}[b]{0.42\textwidth}
  \includegraphics[width=\linewidth]{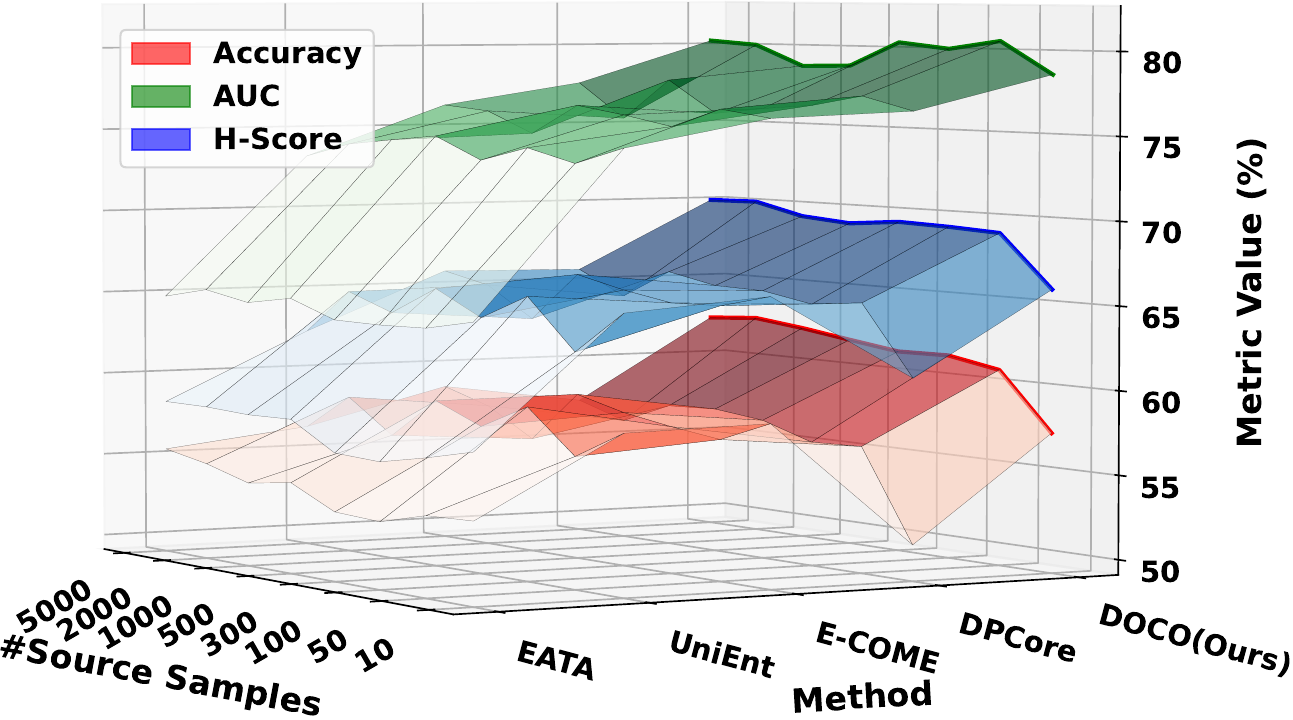}
  \caption{$\#$ Source Sample}
  \label{fig:srcsurface}
\end{subfigure}
\caption{(a) The parameter sensitivity analysis of $\beta$ vs $L$ on H-score (\%). (b) The sensitivity of batch size on H-score (\%). (c) Comparison of five methods with preliminary source characteristics extraction on three metrics versus the number of unlabeled source samples.}
\label{fig:abla}
\end{figure*}

\paragraph{Components Ablation Study.}
We conduct a comprehensive ablation study to dissect the contributions of DOCO's core components: Sample Splitting (\textbf{S}), OOD Propagation (\textbf{O}), and the Structural Regularizer (\textbf{R}). The results in \cref{tab:ablation_compact_fixed_v2} validate that our proposed mechanisms offer broad benefits, enhancing not only the prompt-based (\textbf{P}) approach but also the NormLayer-based (\textbf{N}) baseline, indicating the applicability to the ResNet backbone. A vanilla implementation that relies solely on statistical alignment (P w/o S.O.R) is insufficient, yielding a limited performance gain of $+7.6\%$. DOCO's performance is achieved by building upon this foundation with its unique components. The synergistic application of \textbf{Sample Splitting (S)} and \textbf{OOD Propagation (O)} mechanisms propels the gain substantially to $\bm{+12.1\%}$. The final addition of the \textbf{Structural Regularizer (R)} further refines the feature space, culminating in the full DOCO model's H-score of $70.1\%$. Our analysis makes it clear that DOCO’s success is not merely a consequence of its adaptation paradigm, but is fundamentally driven by its specialized components, which are essential for unlocking robust performance in the demanding OCTTA setting. Notably, even when we disable the splitting mechanism (w/o S.O), the prompt still encodes a reasonable domain pattern on the mixed, noisy ID/OOD batch. This can be attributed to the structural regularizer, which prevents the prompt from aligning OOD semantics to source ID (w/o S.O.R → w/o S.O. yields an additional $\bm{+3.6\%}$). This further addresses a natural concern: even in extreme cases where we cannot clearly split the ID subset, the model does not collapse.

\paragraph{Sensitivity to prompt number $L$ and regularization $\beta$.}
\Cref{fig:heatmap} exhibits a broad plateau on the ImageNet-C and Places365-C combination: DOCO maintains high H-score across a wide swath of $\beta$ and $L$, indicating low sensitivity to precise hyperparameter choices. Notably, along the diagonal trend where $L$ increases, adaptation becomes harder to drive, as excess prompts diffuse learning and weaken domain compensation, but raising $\beta$ counteracts this dispersion, preserving feature geometry and recovering accuracy.

\begin{figure}[t]
\centering
\includegraphics[width=1.0\columnwidth]{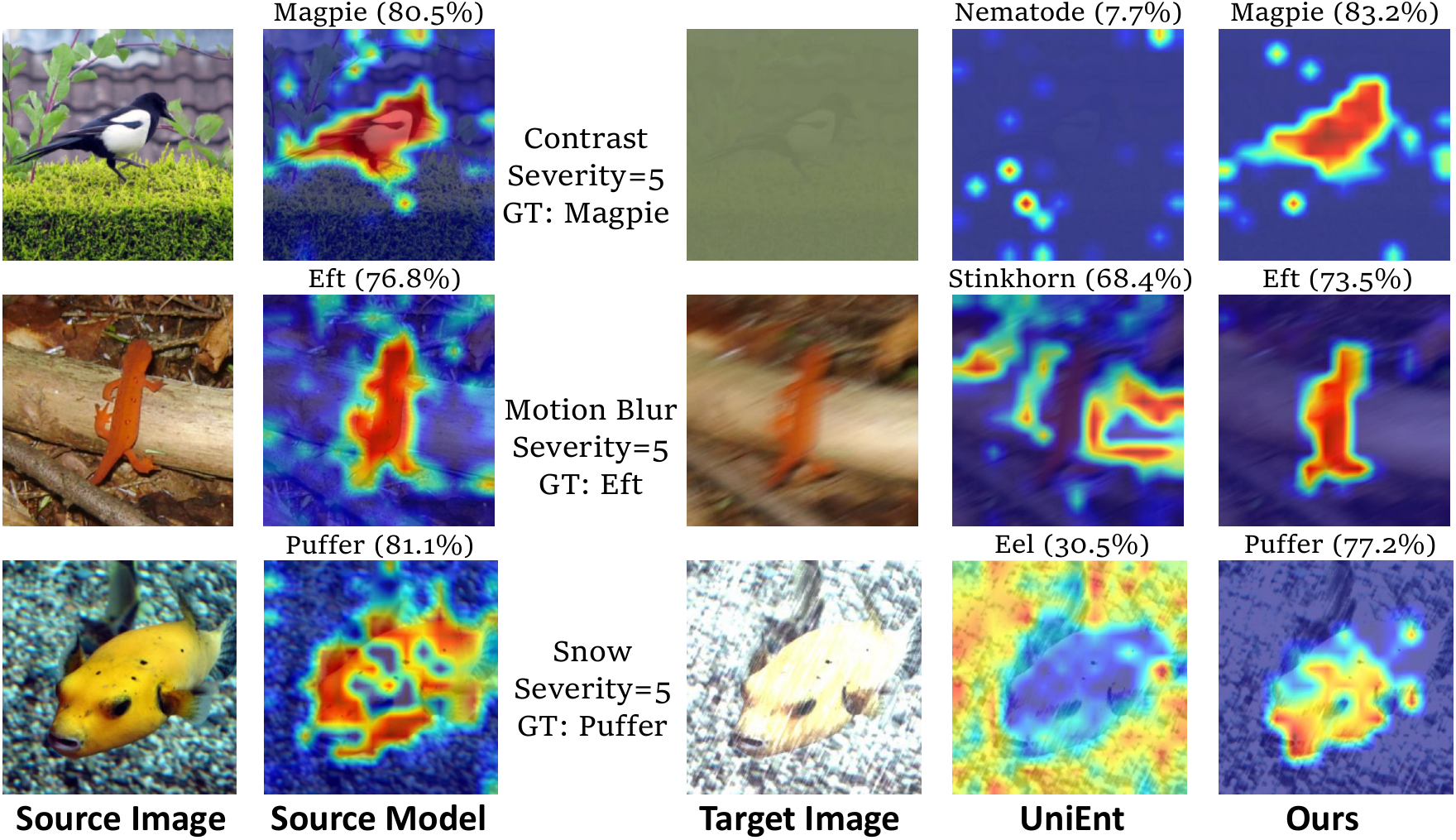} 
\vspace{-10pt}
\caption{Grad-CAM visualizations for the source model on clean images and for UniEnt and DOCO on their corrupted counterparts.}
\label{fig7:gradcam}
\end{figure}

\vspace{-10pt}
\paragraph{Effect on batch size and source number.} 
As shown in \cref{fig:bsline}, DOCO reaches first-tier performance beginning at batch size 8 and remains strong and stable thereafter. In the small-batch regime, removing the structural regularizer noticeably hurts performance relative to the full method, which matches the design insight in \cref{sec:back_learning} when per-batch statistics are noisy. For source number in \cref{fig:srcsurface}, we compare five methods that leverage source number information. Across all source numbers, DOCO consistently attains the highest performance, evidencing high data efficiency. Remarkably, using only 50 source samples already delivers performance very close to the full setting.
For details on the missing points and our small-batch solution of DOCO, please see Appendix~\ref{sec:sup_bsrc}.

\vspace{-10pt}
\paragraph{Visualization.}

As illustrated in \cref{fig7:gradcam}, while the source model focuses on the target object in the clean domain, DOCO better recovers source-like attention under severe corruptions compared to UniEnt \cite{gao2024unified}, concentrating on the true object regions rather than background noise. This qualitative comparison highlights DOCO's core capability: it disentangles domain shifts from semantic features, effectively compensating for the corruption and restoring source-like representations for reliable inference.

\section{Conclusion}

In this paper, we introduce OCTTA, a challenging yet pragmatic setting involving coupled domain and semantic shifts. To address this, we propose DOCO, a novel and efficient framework based on prompt learning. DOCO works by learning to compensate for domain shifts using likely ID data and then propagating this knowledge to enhance the detection of OOD samples, all within a self-reinforcing virtuous cycle. This synergistic design allows DOCO to remain robust and computationally practical where other methods falter. Extensive experiments on multiple challenging benchmarks validate the superiority of our method, which sets a new SOTA in handling coupled shifts.

\section*{Acknowledgment}

This work was supported in part by National Key R\&D Program of China (2024YFB3908500, 2024YFB3908502), NSFC Excellent Young Scientists Fund Program (Overseas), Guangdong Basic and Applied Basic Research Foundation (2023B1515120026), Shenzhen Science and Technology Program (KJZD20240903100022028), and Scientific Development Funds from Shenzhen University.
{
    \small
    \bibliographystyle{ieeenat_fullname}
    \bibliography{main}
}

\appendix
\clearpage
\setcounter{page}{1}
\maketitlesupplementary

In this appendix, we provide detailed supplementary materials to further clarify and support our framework. We begin with additional analysis of DOCO, where we present the full algorithmic procedure, discuss its connection to domain compensation and feature disentanglement, and examine how the learned prompts generalize to unseen domains with extended visualizations. We then describe implementation details, including the construction of corrupted datasets, the configurations of all baselines, and practical considerations such as batch-size stabilizers and the use of source-domain samples. Finally, we report extended experimental results, covering computational efficiency, robustness under different OOD ratios and corruption severities, and comprehensive comparisons across multiple OOD score measurements to validate the stability and effectiveness of DOCO in the OCTTA setting.

\section{Additional Analysis of DOCO}
\subsection{Algorithm}
\label{sec:sup_alg}

\begin{algorithm}[h]
\caption{DOmain COmpensation (DOCO)}
\label{alg:doco}
\small
\begin{algorithmic}[1]
\REQUIRE Model $f_\theta=h\circ\phi$; source labels $\mathcal{Y}^S$; cached source statistics $(\mu_S,\sigma_S)$; frozen classifier weights $\{w_c\}_{c\in\mathcal{Y}^S}$; test stream $\{\mathcal{B}_t\}_{t=1}^{T}$; learning rate $\eta$; regularization weight $\beta$.
\ENSURE Predictions on all batches and updated prompts $\{p_t\}$.

\STATE Initialize the first prompt $p_1$ with Xavier-uniform initialization.

\STATE \textbf{First-batch initialization ($t=1$):}
\STATE Compute raw features $Z_{1,\mathrm{raw}}=\{\phi(x)\}_{x\in\mathcal{B}_1}$.
\STATE Compute raw prototypical distances by Eq.~\eqref{eq:proto-distance} on $Z_{1,\mathrm{raw}}$.
\STATE Run $K$-Means ($K=2$) on the raw score set $\mathcal{S}_1^{\mathrm{raw}}=\{d_{\mathrm{proto}}(\phi(x))\mid x\in\mathcal{B}_1\}$, and split $\mathcal{B}_1$ into $\hat{\mathcal{B}}_1^{\mathrm{ID}}$ and $\hat{\mathcal{B}}_1^{\mathrm{OOD}}$ analogously to Eq.~\eqref{eq:split-compact}.
\STATE Predict samples in $\hat{\mathcal{B}}_1^{\mathrm{ID}}$ using the raw model $h(\phi(x))$.
\FOR{$k=1$ to $50$}
    \STATE Update $p_1$ on $\hat{\mathcal{B}}_1^{\mathrm{ID}}$ by minimizing $\mathcal{L}_{\mathrm{DOCO}}$ in Eq.~\eqref{eq:L_DOCO}.
\ENDFOR
\STATE Set $p_2\leftarrow p_1$.
\STATE Predict samples in $\hat{\mathcal{B}}_1^{\mathrm{OOD}}$ by Eq.~\eqref{eq:ood_prediction} using $p_2$.

\FOR{$t=2$ to $T$}
    \STATE Compute prompted features $Z_{t,p}=\{\phi(x;p_t)\}_{x\in\mathcal{B}_t}$.
    \STATE Compute prototypical distances by Eq.~\eqref{eq:proto-distance} on $Z_{t,p}$.
    \STATE Run $K$-Means ($K=2$) on $\mathcal{S}_t=\{d_{\mathrm{proto}}(z)\mid z\in Z_{t,p}\}$, and obtain $\hat{\mathcal{B}}_t^{\mathrm{ID}}$ and $\hat{\mathcal{B}}_t^{\mathrm{OOD}}$ by Eq.~\eqref{eq:split-compact}.
    \STATE Predict samples in $\hat{\mathcal{B}}_t^{\mathrm{ID}}$ using the current prompt $p_t$.
    \STATE Update the prompt on $\hat{\mathcal{B}}_t^{\mathrm{ID}}$ by one gradient step on Eq.~\eqref{eq:L_DOCO} to obtain $p_{t+1}$.
    \STATE Predict samples in $\hat{\mathcal{B}}_t^{\mathrm{OOD}}$ by Eq.~\eqref{eq:ood_prediction} using $p_{t+1}$.
\ENDFOR
\end{algorithmic}
\end{algorithm}

As mentioned in \emph{Implementation Details}, the prompts conduct a one-time self-supervised update for 50 iterations to refine their initial state. For all subsequent batches we reuse the prompt and perform only a single gradient step.

\subsection{End-to-End Domain Compensation}
\label{sec:sup_three-pathways-compact}

\noindent\textbf{Two feature-level routes.}
Pixel-space restoration ($g^{-1}$) could in principle clean inputs before feature extraction, but its ill-posedness risks artifacts propagating to features; we therefore focus on \emph{feature-level} compensation.
A representative \emph{explicit separation} route is DICS~\cite{miao2025dics}: during \emph{training}, it learns domain vectors and subtracts them while enforcing same-class cross-domain consistency (DIT), and further promotes class specificity via a memory-driven soft labeling (CST), then \emph{deploys a fixed model} without using the target stream.
In contrast, our route performs \emph{test-time, in-process} correction: within each batch $t$, we estimate the shared factor $\delta_t$ using only likely ID samples, and immediately propagate the learned prompt $p_{t+1}$ to the whole batch during the forward pass,
yielding $\phi(x;p_{t+1}) \approx \phi(x)-\delta_t \approx s(x)$ for both ID and OOD candidates from the same batch. This \emph{batch-consistent} compensation leverages the live stream \emph{inside} the feature extractor and avoids back-propagating through likely OOD samples.

\begin{figure}[h]
    \centering
    \includegraphics[width=0.6\columnwidth]{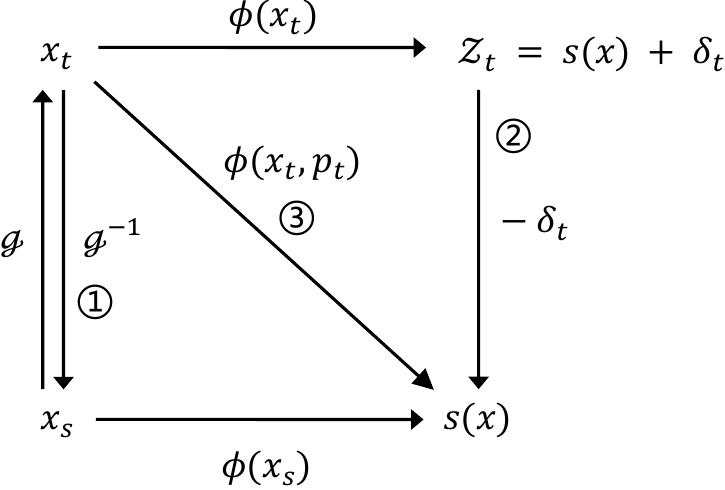}
    \caption{Three pathways from a corrupted image $x_t$ to a domain-invariant feature $s(x)$:
    (1) pixel restoration (briefly noted), (2) \emph{training-time} explicit separation (DICS), and (3) our \emph{test-time} in-process correction that learns a batch-conditioned prompt inside $\phi(\cdot,p)$.}
    \label{fig:3pathways}
\end{figure}

\noindent\textbf{Relation to DICS.}
Both routes aim to expose $s(x)$ by attenuating domain factors. Empirically, DICS realizes this via \emph{training-time} explicit subtraction plus class-specific constraints, whereas we realize a \emph{test-time} compensation conditioned on the current batch and updated online \emph{without} backprop on likely-OOD data—thereby preventing OOD semantics from contaminating alignment and stabilizing the decision boundary under a frozen head.

\subsection{Generalization to Unseen Domains}
\label{sec:sup_prompt-generalization}

We evaluate whether the learned prompt generalizes across unseen domains \emph{before} any update on the new domain. 
For each domain transition in a sequence, we take the very first target batch (except the first domain for which the prompt is initialized) and compute the statistical misalignment $\mathcal{L}_{stat}$ against pre-cached source statistics. 
We compare (i) the static \textbf{Source} model and (ii) \textbf{DOCO} carrying the prompt updated on \emph{previous} domains but untouched on the current one. 
On both ImageNet-C and LAION-C streams (see \cref{fig:sup_6orderC} and \cref{fig:sup_6orderL}; six random orders are examined for each), DOCO consistently exhibits a lower initial $\mathcal{L}_{stat}$, indicating a zero-backprop corrective effect that transfers to novel domains. 
In a few difficult transitions the initial gap is small, yet the loss still decreases rapidly without degradation, suggesting the prompt provides a beneficial starting point rather than causing negative transfer.

In short, the prompt functions as a batch-wise domain compensator that generalizes to new domains at encounter time, aligning features toward the source geometry and enabling stable adaptation in OCTTA.

\subsection{Extended visualization results.}
\label{sec：sup_easyhard}

\begin{figure}[h]
\centering
\includegraphics[width=1.0\columnwidth]{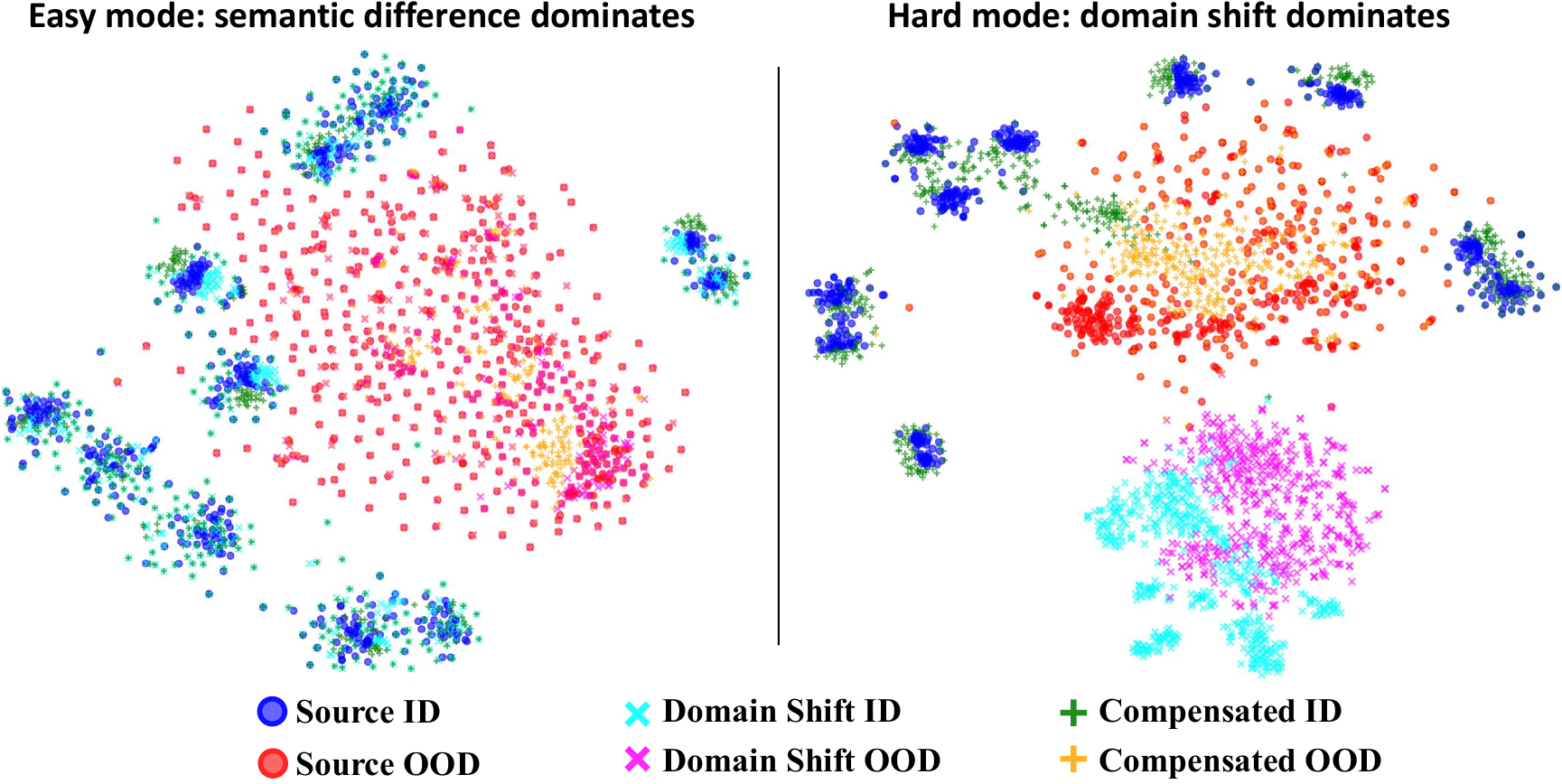} 
\caption{Semantic and domain shift antagonism}
\label{fig4:easyvshard}
\end{figure}
To further visualize this internal mechanism, we present a t-SNE \cite{maaten2008visualizing} visualization of the feature space in \cref{fig4:easyvshard}, contrasting a mild domain shift (brightness) with a severe one (fog). In the \textbf{Hard mode}, the severe domain shift overwhelms the semantic differences. This causes the features of both shifted ID (\textcolor{shiftidlightblue}{$\times$}) and OOD (\textcolor{shiftoodpink}{$\times$}) samples to drift significantly from their origins and mix together in the feature space. Crucially, DOCO effectively reverses this effect, pulling the compensated features (\textcolor{compensatedidgreen}{$+$} and \textcolor{compensatedoodyellow}{$+$}) back to align with their corresponding source ID (\textcolor{blue}{$\bullet$}) and OOD (\textcolor{red}{$\bullet$}) clusters. Conversely, in the \textbf{Easy mode}, the intrinsic semantic differences dominate the mild domain shift. Here, aided by our pairwise structural regularizer, DOCO demonstrates its precision by ensuring the compensated features (\textcolor{compensatedidgreen}{$+$} and \textcolor{compensatedoodyellow}{$+$}) remain tightly anchored to their respective clusters (\textcolor{blue}{$\bullet$} and \textcolor{red}{$\bullet$}) without introducing distortion. This confirms DOCO's dual ability to robustly correct large shifts while delicately preserving feature structures under smaller ones.

\begin{figure}[t] 
  \centering

  \newlength{\imgsep}\setlength{\imgsep}{2pt}     
  \newlength{\innerpad}\setlength{\innerpad}{0pt}

  \newlength{\oneimgH}
  \setlength{\oneimgH}{%
    \dimexpr(\textheight - \abovecaptionskip - \belowcaptionskip
    - 5\imgsep - 2\innerpad)/6\relax}

  \vspace*{\innerpad}
  \includegraphics[width=\linewidth,height=\oneimgH]{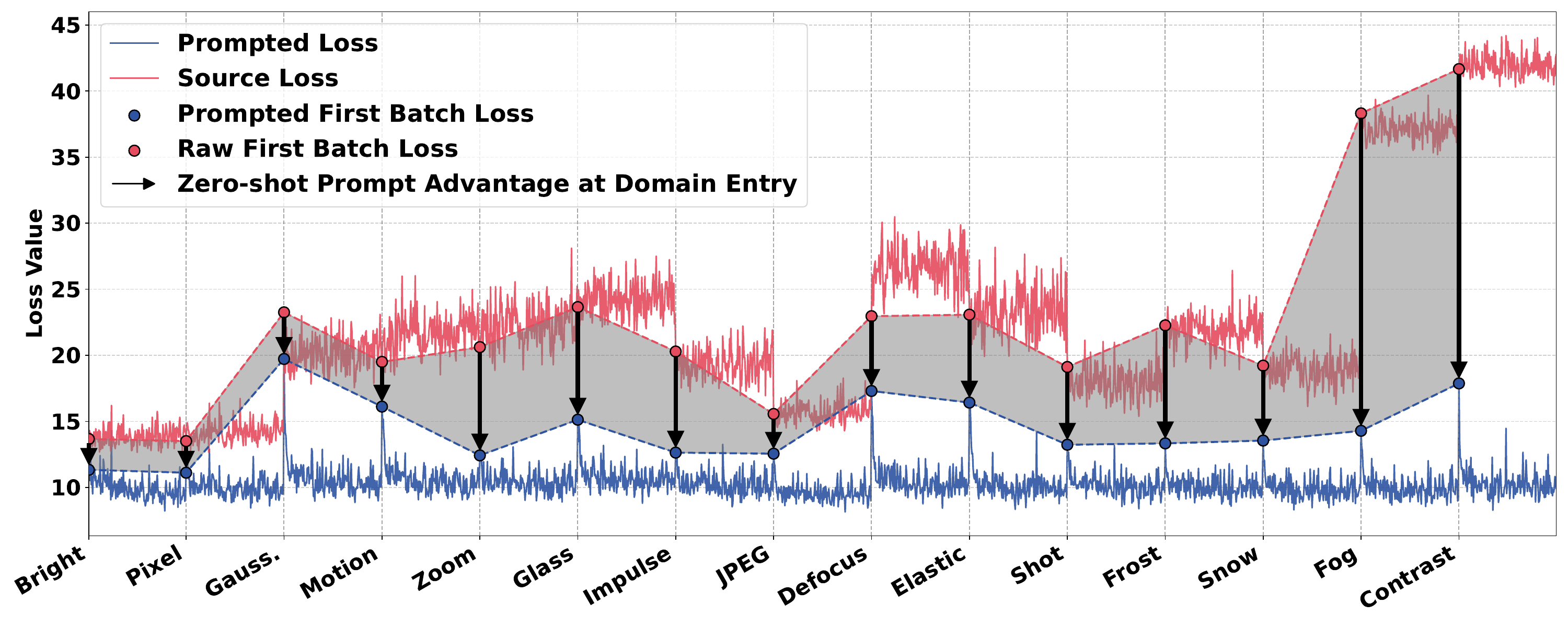}\vspace{\imgsep}
  \includegraphics[width=\linewidth,height=\oneimgH]{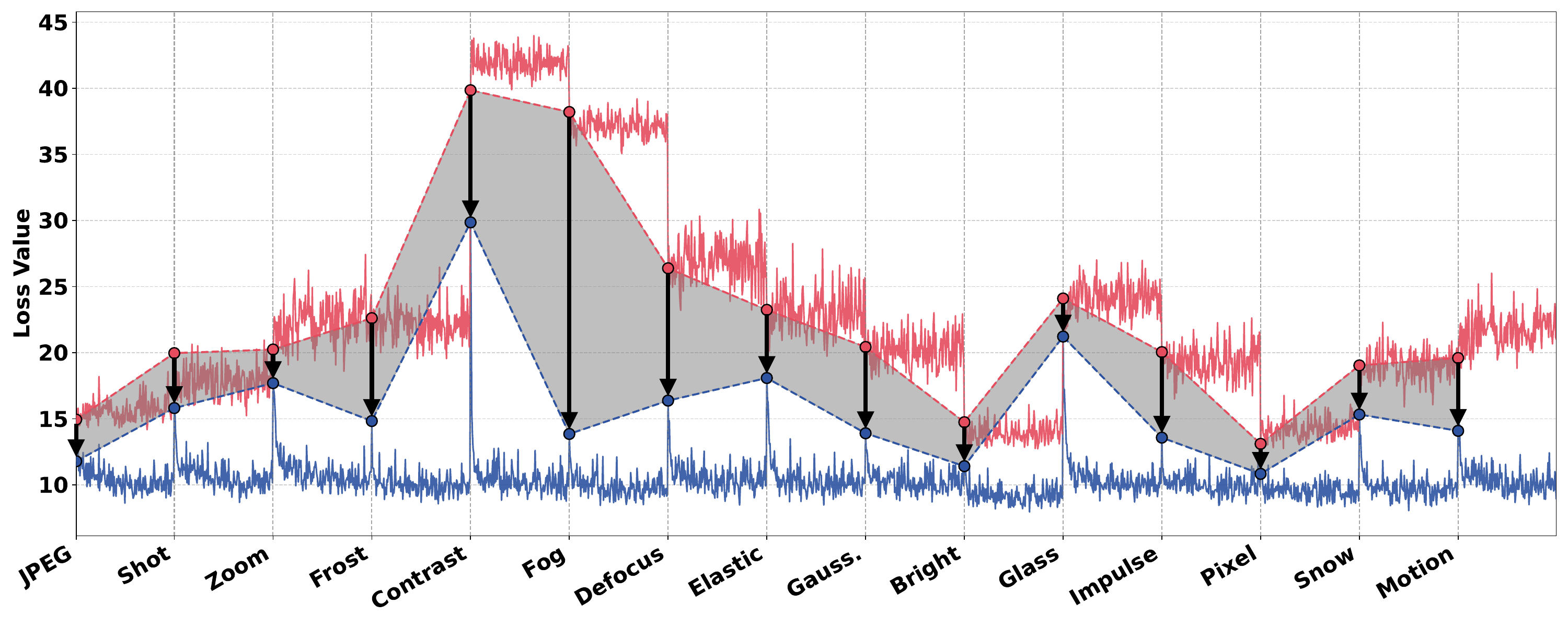}\vspace{\imgsep}
  \includegraphics[width=\linewidth,height=\oneimgH]{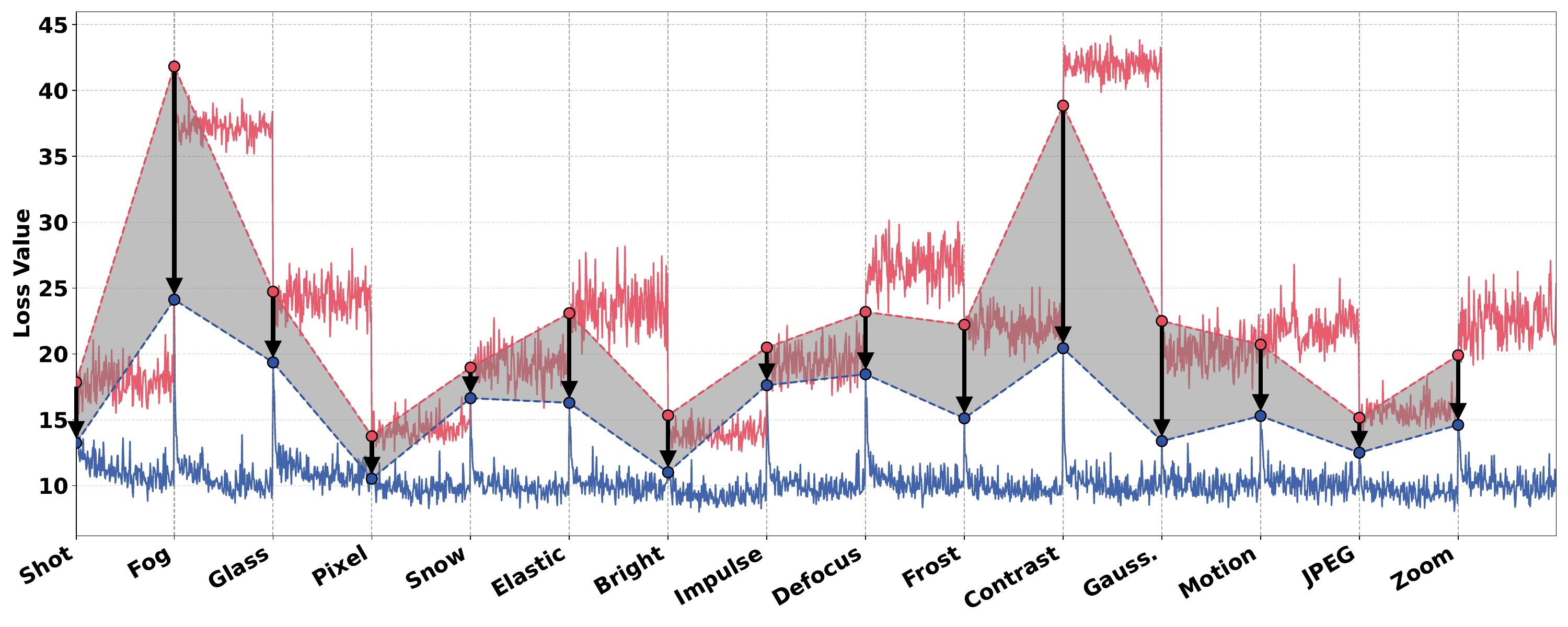}\vspace{\imgsep}
  \includegraphics[width=\linewidth,height=\oneimgH]{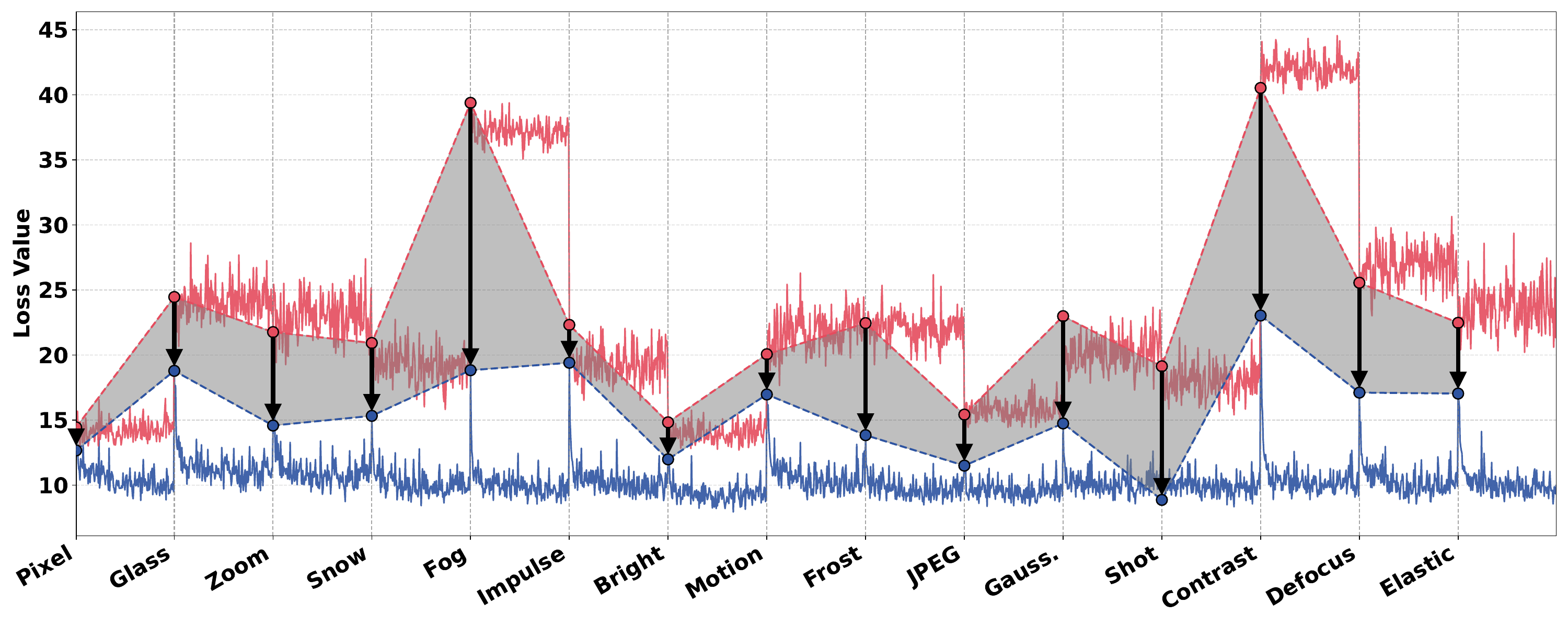}\vspace{\imgsep}
  \includegraphics[width=\linewidth,height=\oneimgH]{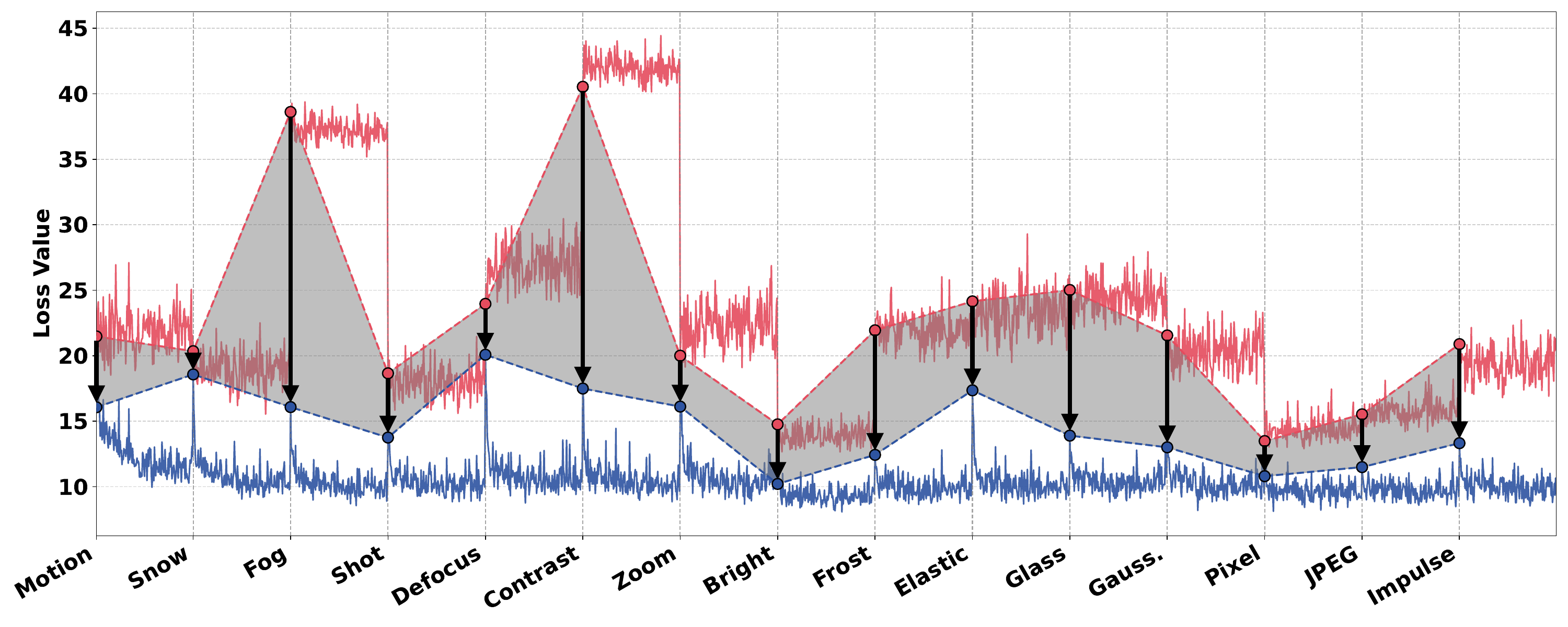}\vspace{\imgsep}
  \includegraphics[width=\linewidth,height=\oneimgH]{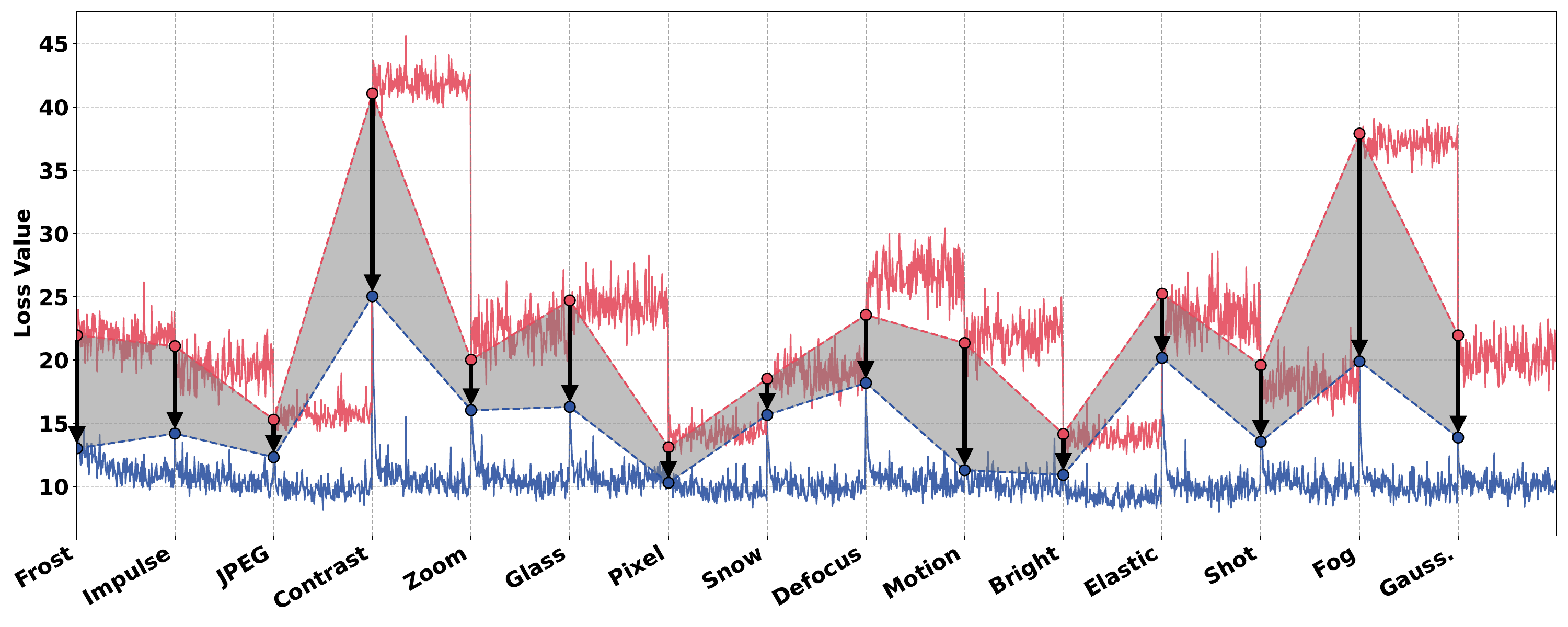}
  \vspace*{\innerpad}

  \caption{First-batch statistical loss per domain in six different OCTTA orders (ImageNet-C, $\kappa=0.5$).}
  \label{fig:sup_6orderC}
\end{figure}

\begin{figure}[t] 
  \centering

  \setlength{\oneimgH}{%
    \dimexpr(\textheight - \abovecaptionskip - \belowcaptionskip
    - 5\imgsep - 2\innerpad)/6\relax}

  \vspace*{\innerpad}
  \includegraphics[width=\linewidth,height=\oneimgH]{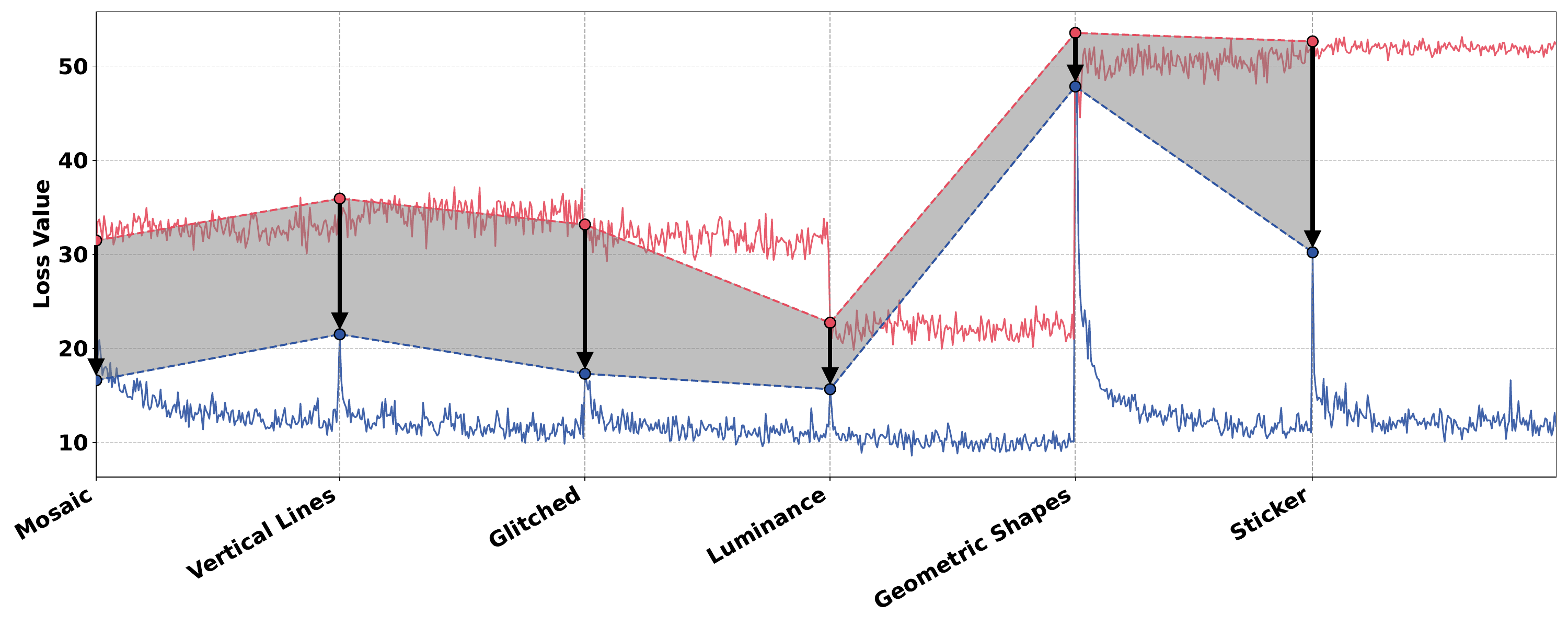}\vspace{\imgsep}
  \includegraphics[width=\linewidth,height=\oneimgH]{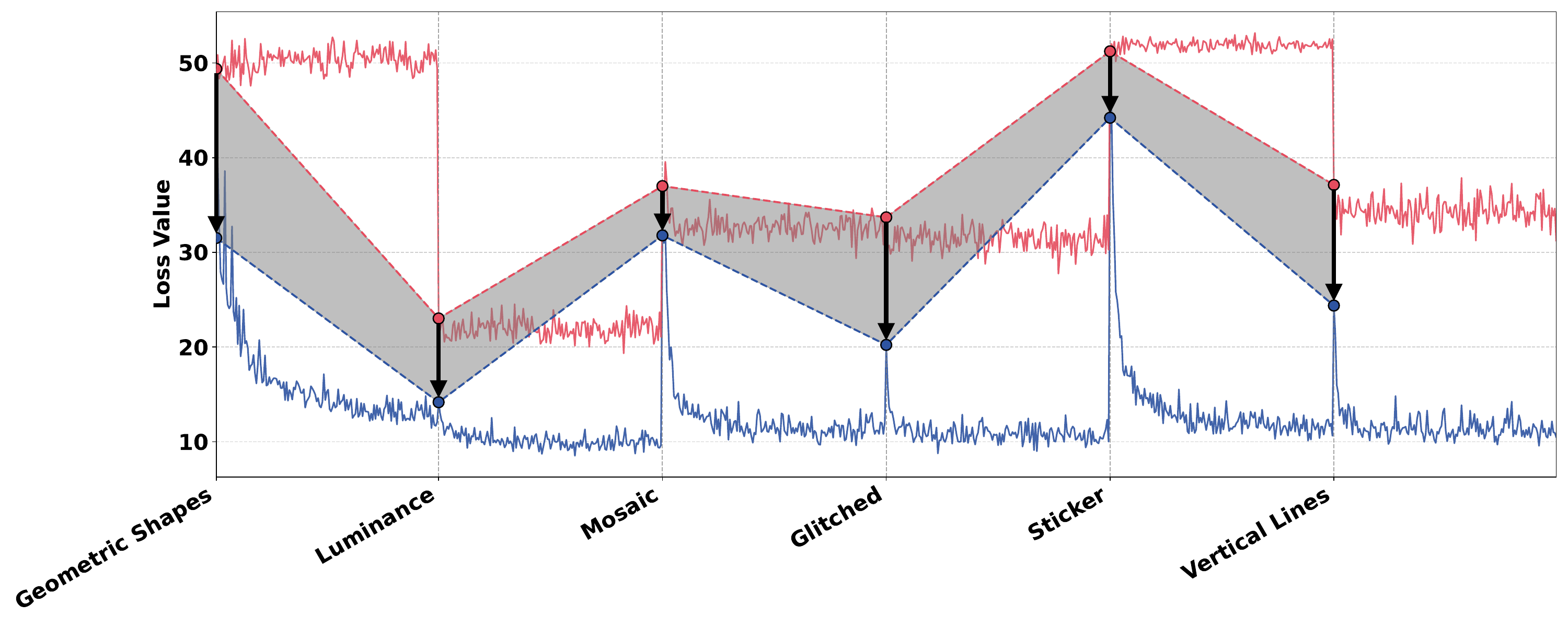}\vspace{\imgsep}
  \includegraphics[width=\linewidth,height=\oneimgH]{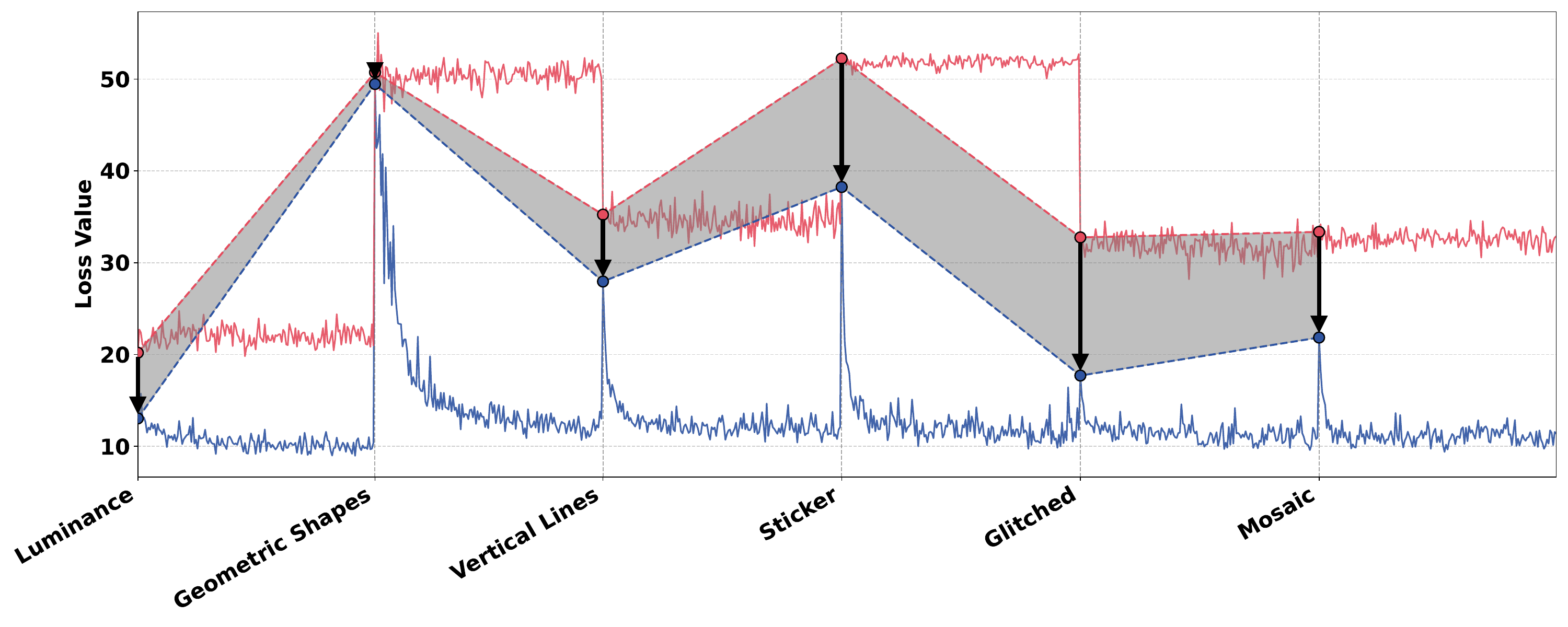}\vspace{\imgsep}
  \includegraphics[width=\linewidth,height=\oneimgH]{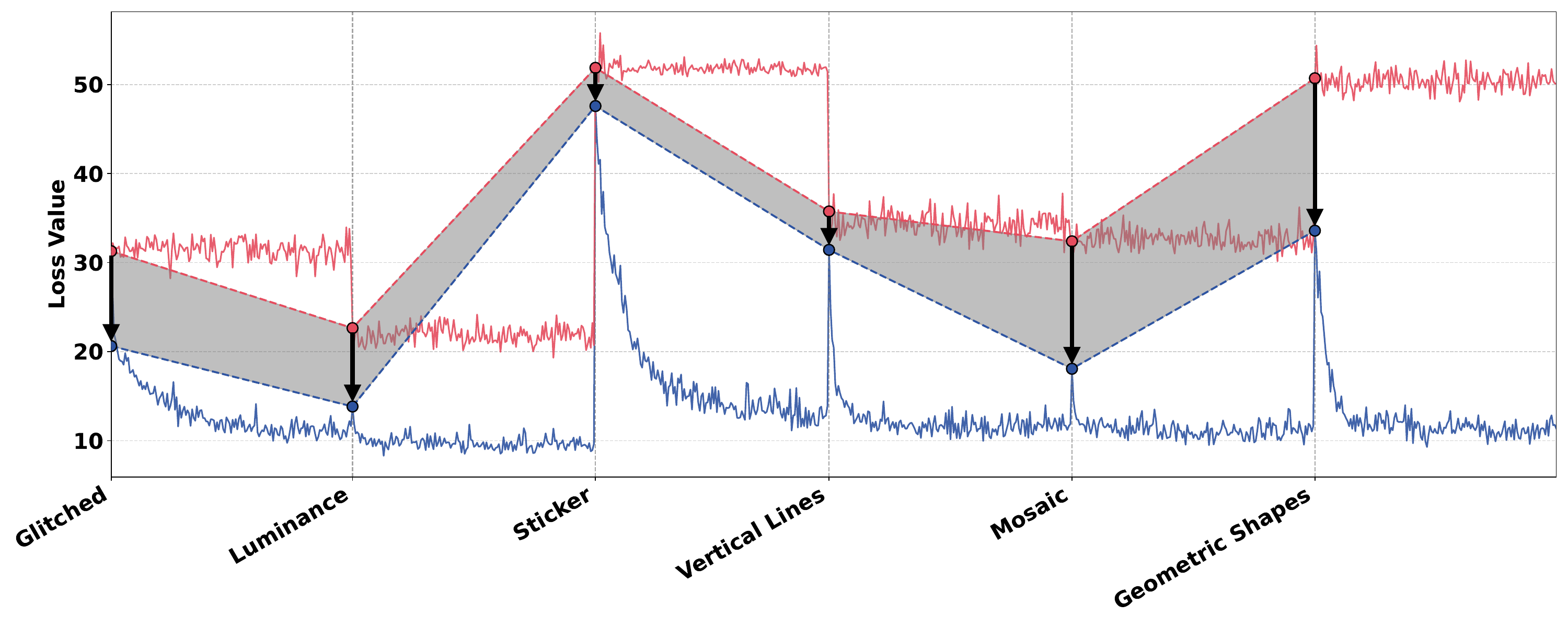}\vspace{\imgsep}
  \includegraphics[width=\linewidth,height=\oneimgH]{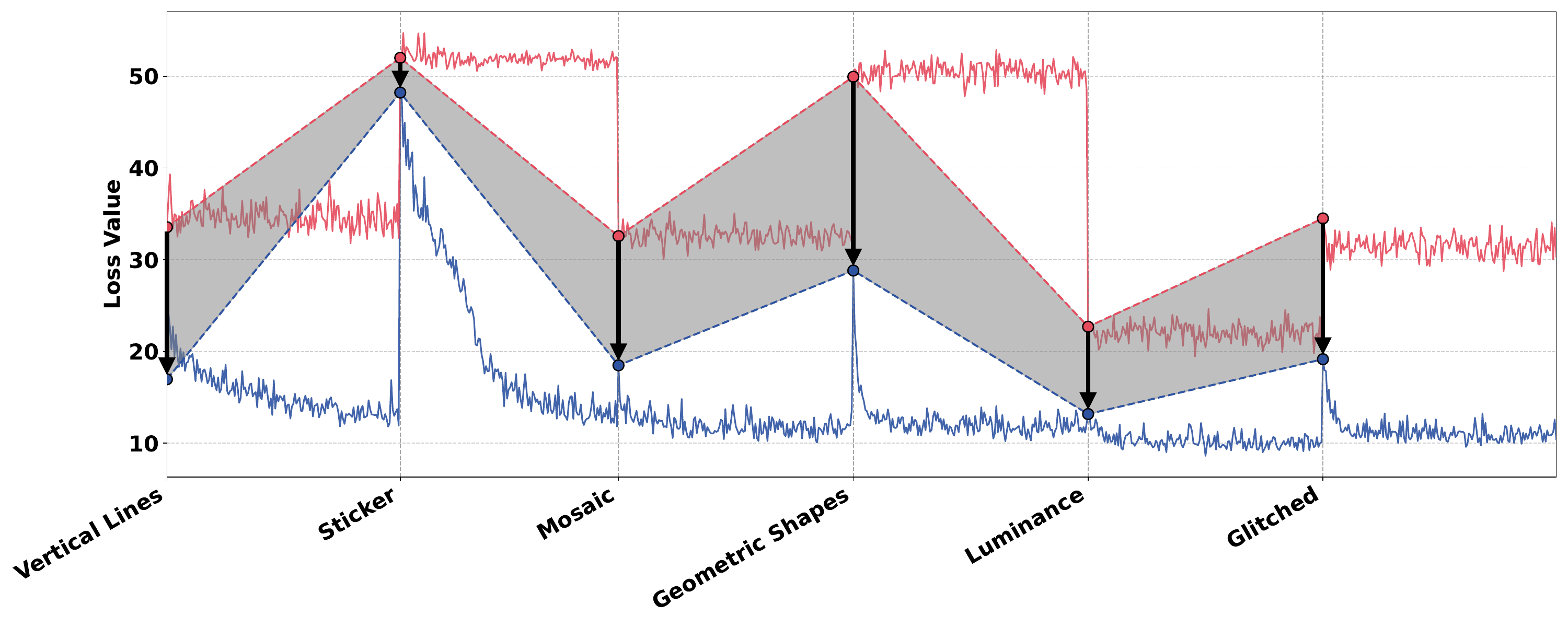}\vspace{\imgsep}
  \includegraphics[width=\linewidth,height=\oneimgH]{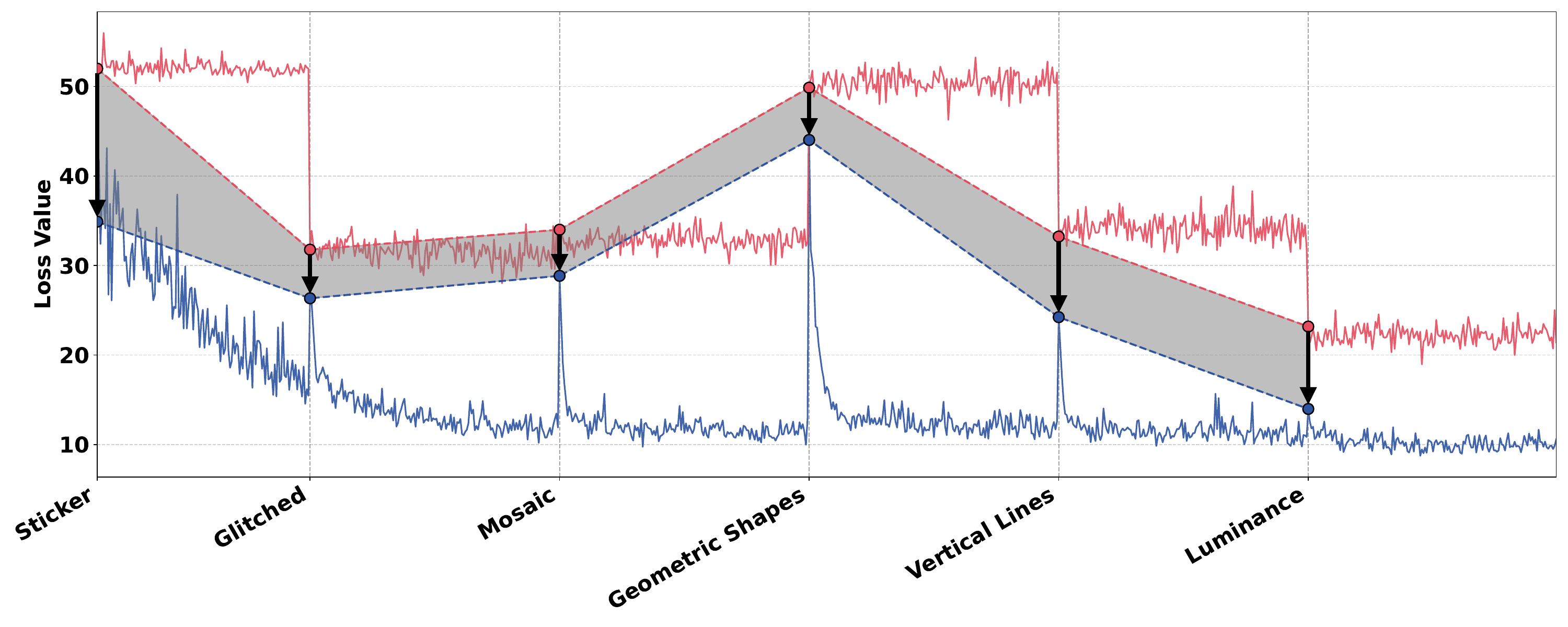}
  \vspace*{\innerpad}

  \caption{First-batch statistical loss per domain in six different OCTTA orders (LAION-C, $\kappa=0.5$).}
  \label{fig:sup_6orderL}
\end{figure}

\section{Implementation Details}

\begin{figure*}[h]
\centering
\includegraphics[width=1\textwidth]{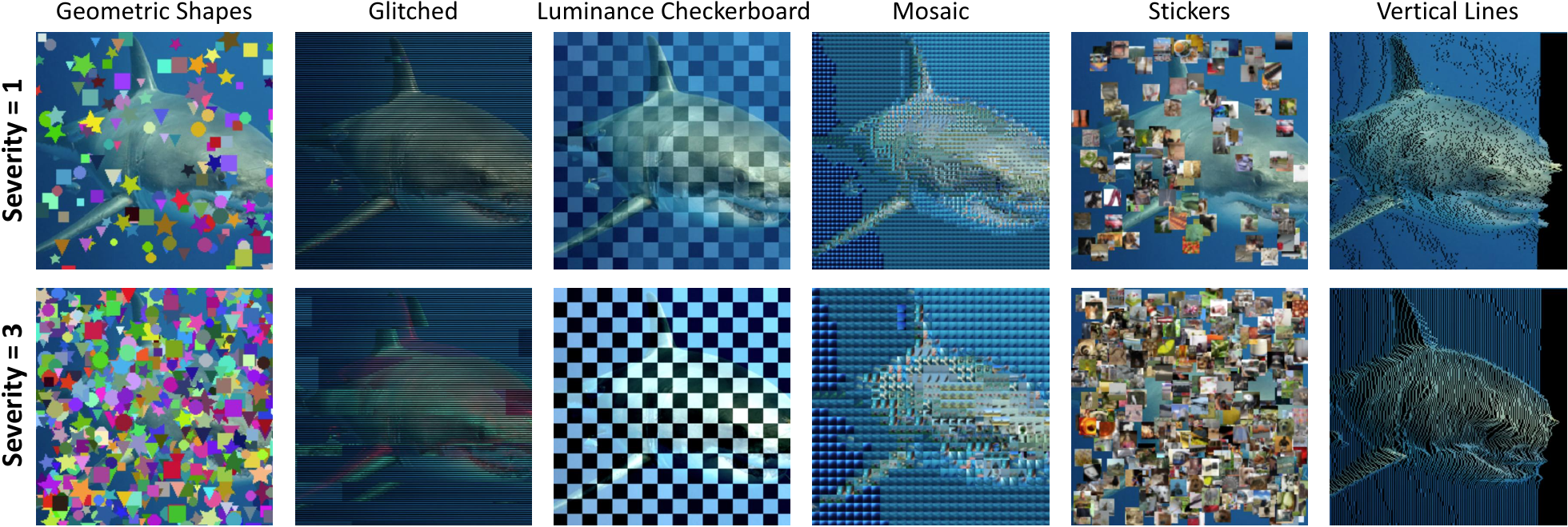}
\caption{ Examples of six LAION-C corruption types.}
\label{fig:LAION_example}
\end{figure*}

\subsection{Dataset Corruption Settings.}
\label{sec:sup_laion}
The LAION-C benchmark features six highly challenging domains: \textit{Mosaic, Glitched, Vertical Lines, Geometric Shapes, Stickers,} and \textit{Luminance Checkerboard}. Example corruptions are shown in \cref{fig:LAION_example}. To evaluate covariate-shifted OOD robustness, we applied these synthetic corruptions to OOD datasets. The specific generation settings for \texttt{mosaic} and \texttt{sticker} are as follows:
\begin{itemize}
    \item \textbf{Tile Pool Source}: We used the ImageNet-1K (\texttt{ILSVRC2012}) validation set as the tile pool for corruption generation due to its diversity.
    \item \textbf{Tile Pool Subsampling}: To manage memory constraints, we subsampled 5000 images from the 50,000-image validation set, sequentially selected and packaged into a \texttt{.tar} archive using the \texttt{WebDataset} format, as required by the LAION-C data loader.
    \item \textbf{Corruption Generation}: Corruptions were applied using the curated 5000-image tile pool. All parameters, such as \texttt{intensity\_level}, and generation protocols followed the default behavior of the LAION-C codebase, except for the specified sub-sampling strategy.
\end{itemize}
We use a fixed random seed when subsampling the 5,000 validation images and generating corruptions to ensure that LAION-C benchmarks are fully reproducible.

\subsection{Baselines.}
\label{sec:sup_baseline}

For a fair and reproducible comparison, we implement all baseline methods using their official, publicly available codebases. We initialize all hyperparameters and learning rates for each algorithm strictly according to the configurations recommended by the original authors. The implementation of OSTTA is taken from the official UniEnt repository. For the ViDA baseline, we evaluate two backbone settings: (i) a standard pre-trained model from the \texttt{timm} library, and (ii) the pre-trained model released in the official ViDA repository. In the latter case, the low-rank and high-rank ViDA modules are pre-trained, providing a much better initialization than random parameters. In our final results, we report the performance of the ViDA variant that achieves the higher score between these two configurations. We consider both backbone settings to avoid penalizing ViDA due to implementation differences, and always report the better one, while all other baselines are evaluated with their official configurations. For STAMP, we follow its experimental setup on the ImageNet benchmark and remove the consistency filtering mechanism to avoid discarding too many samples. For E-COME, UniEnt, EATA, and DPCore, which require collecting information from the source domain beforehand, we also adhere to their default settings. Specifically, for E-COME, UniEnt, and EATA, the number of source samples used to compute the Fisher information matrix is set to 2000, while DPCore uses 300 source samples by default. To ensure a fair comparison, DOCO is likewise restricted to only 300 source-domain samples. Moreover, to keep the number of parameters comparable, although DOCO can obtain better performance with a larger prompt number (\cref{fig:heatmap}), in all main experiments we fix the prompt length to $L=8$, which matches the configuration used by DPCore.

\subsection{Details for Batch Size and Source Number}
\label{sec:sup_bsrc}

\paragraph{Note on omitted points.} We provide the data integrity explanation of \cref{fig:bsline} here for further understanding. \textbf{DPCore} at small batches (BS=2/4/8) on mixed data exhibits core-set blowup due to unstable per-batch statistics, rendering runs infeasible. \textbf{EATA-based} variants (EATA, E-COME, UniEnt) at BS=128 are omitted because their GPU memory cost is prohibitive.

\vspace{-10pt}
\paragraph{Small-batch stabilizers in DOCO.} In the small-batch regime (test batch size $\leq 8$) discussed in \S\emph{Effect on batch size and source number}, DOCO enables two lightweight stabilizers that are only activated in this analysis and are disabled in all main results. (1) A FIFO buffer $\mathcal{R}$ of a fixed size (we use $64$ recent values in our experiments) stores recent proto-distances $d_{\mathrm{proto}}(z)$; we run $k$-means ($K{=}2$) over scores in $\mathcal{R}$ and use the resulting clusters to assign the current batch to ID/OOD, reducing the variance of the split when batches are tiny. (2) We enforce a minimum of one ID sample to update $p_t$; otherwise, we skip adaptation and only perform forward prediction. The structure-preserving regularizer $\mathcal{L}_{\mathrm{reg}}$ is the Frobenius norm between the pairwise cosine-similarity matrices of prompted and raw CLS features, and is evaluated only when the ID subset has at least two samples.

\begin{figure*}[t]
\centering 
\begin{subfigure}[b]{0.33\textwidth} 
    \centering
    \includegraphics[width=\linewidth]{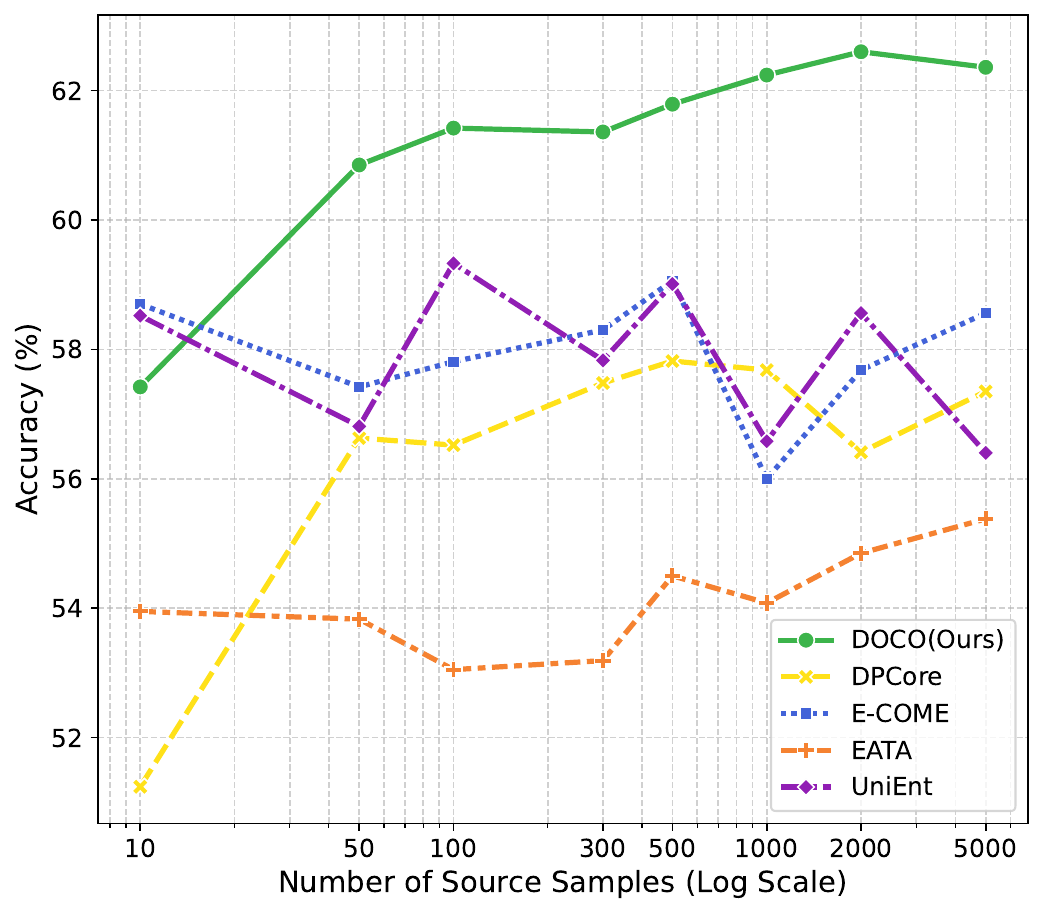}
    \caption{Accuracy(\%)}
    \label{fig:sup_srcacc}
\end{subfigure}
\begin{subfigure}[b]{0.33\textwidth} 
    \centering
    \includegraphics[width=\linewidth]{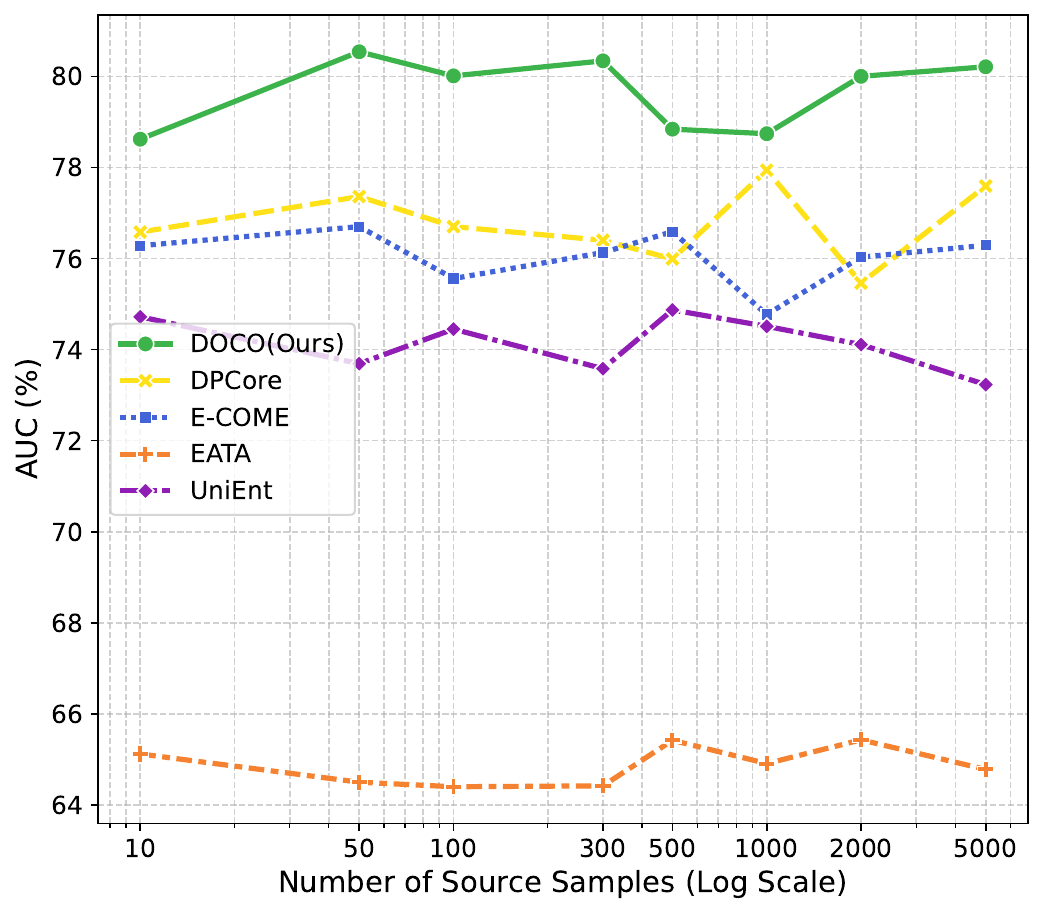} 
    \caption{AUC(\%)}
    \label{fig:sup_srcauc}
\end{subfigure}
\begin{subfigure}[b]{0.33\textwidth} 
    \centering
    \includegraphics[width=\linewidth]{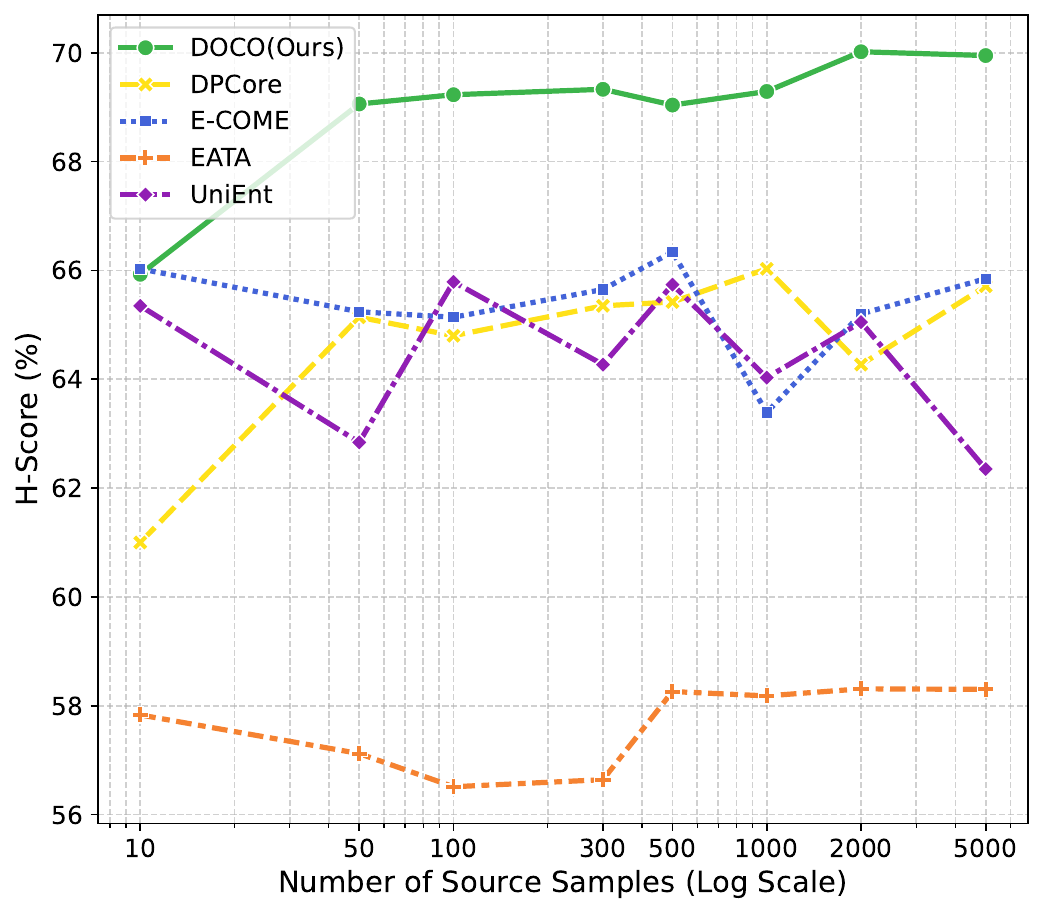} %
    \caption{H-score(\%)}
    \label{fig:sup_srchs} 
\end{subfigure}
\caption{\textbf{Source-number and small-batch ablations.}
(a)–(c): Accuracy, AUC, and H-score vs. number of source samples (log scale).}
\label{fig:src_batch_appendix}
\end{figure*}

\noindent\textbf{More details on source number ablation.}
As a supplementary view of \cref{fig:srcsurface}, we additionally provide a 2D source number comparison visualization on \cref{fig:src_batch_appendix}.

\section{Extended Experimental Results}

\subsection{Computational efficiency}

\begin{table}[t]
  \centering
  \small 
  \setlength{\tabcolsep}{3.0pt} 
  \caption{Comparison w.r.t computational complexity}
  \label{tab:sup_compute} 
  \renewcommand{\arraystretch}{0.8} 
  \begin{tabular}{ccccc} 
    \toprule
    Method & Update & Memory(MB) & Time & H-s (\%) \\
    \midrule
    Tent \cite{wang2021tent}       & Norm     & 10,094/10,112 & 1.0/1.0 & 0.6/23.8 \\
    CoTTA \cite{wang2022continual} & All      & 20,780/21,354 & 2.8/4.7 & 18.5/54.8  \\
    EATA \cite{niu2022efficient}   & Norm     & 12,442/12,442 & 0.9/0.84 & 27.9/57.8 \\
    SAR \cite{niu2023towards}      & Norm     & 12,542/6,800 & 1.5/1.6 & 9.6/54.3 \\
    OSTTA \cite{lee2023towards}    & Norm     & 12,944/12,618 & 1.9/1.9 & 17.5/58.5 \\
    ViDA \cite{liu2024vida}        & Adapters & 11,812/11,786 & 8.6/8.7 & 3.9/48.4 \\
    UniEnt \cite{gao2024unified}   & Norm     & 13,944/13,944 & 1.4/1.4 & 29.3/65.4 \\
    STAMP \cite{yu2024stamp}       & Norm     & 9,978/10,070  & 2.8/2.8 & 19.5/60.2 \\
    E-COME \cite{zhang2025come}    & Norm     & 12,494/12,494 & 0.7/0.8 & 19.9/65.2 \\
    S-COME \cite{zhang2025come}    & Norm     & 6,966/6,800  & 1.6/1.6 & 0.3/45.5 \\
    DPCore \cite{zhang2025dpcore}  & Prompts  & 11,424/10,700 & 3.6/2.0 & 30.3/62.6 \\ 
    \rowcolor{blue!10}
    \textbf{DOCO (Ours)}                  & Prompts  & 14,694/16,604 & 2.1/1.9 & 32.7/70.1 \\
    \bottomrule
  \end{tabular}
\end{table}

\cref{tab:sup_compute} reports runtime and memory under the same protocol as the main results: batch size $64$ on \textbf{LAION-C (sev=3)} and \textbf{ImageNet-C (sev=5)}. 
Numbers are shown as ``LAION-C/ImageNet-C''. 
\emph{Time} is a relative measure normalized to Tent$\,{=}\,1.0$ (lower is faster), and \emph{Memory} is GPU usage (MB). All methods are measured on a single NVIDIA Quadro P6000 GPU under the same implementation, so the relative runtime is directly comparable.
Overall, \textbf{DOCO} attains the strongest performance in the main tables while keeping \emph{moderate} overhead—its prompt-based updates add little computation compared to methods that retrain normalization layers or adapters. 
Notably on the harder LAION-C (sev=3), DPCore’s core-set rapidly grows during the stream, inflating computation and wall-clock time. 
These results confirm that our in-process prompt correction offers a favorable accuracy–efficiency trade-off in OCTTA.

\subsection{Different OOD percentage}
\label{appendix:experiment}

We thoroughly analyze the impact of varying OOD sample percentages on model performance, with overview \cref{tab:oodpercent_hs} and detailed results for OOD ratios $10\% - 40\%$ in  \cref{tab_sup:imgNetC_10} - \cref{tab_sup:imgNetC_40}. DOCO demonstrates strong robustness, delivering consistently high accuracy across all tested OOD ratios, and in particular achieves a $\bm{5\%}$ favorable improvement over the next-best method DPCore when $\kappa = 0.4$.

\begin{table*}[t]
\centering
\small
\setlength{\tabcolsep}{2.0pt}
\caption{Results ($\%$) for ImageNet-to-ImageNet-C benchmark (severity = 5) in OCTTA setting with different OOD samples percentages across six covariate-shifted OOD datasets. All the results are averaged over 15 domains. }
\renewcommand{\arraystretch}{0.85}
\begin{tabular}{c | ccc | ccc | ccc | ccc | ccc | ccc}
\toprule
\multirow{2}{*}{\textbf{Method}} &
\multicolumn{3}{c}{\textbf{10\%}} &
\multicolumn{3}{c}{\textbf{20\%}} &
\multicolumn{3}{c}{\textbf{30\%}} &
\multicolumn{3}{c}{\textbf{40\%}} &
\multicolumn{3}{c}{\textbf{50\%}} &
\multicolumn{3}{c}{\textbf{Avg.}} \\
\cmidrule(lr){2-4} \cmidrule(lr){5-7} \cmidrule(lr){8-10} \cmidrule(lr){11-13} \cmidrule(lr){14-16} \cmidrule(lr){17-19}
& ACC & AUC & H-s & ACC & AUC & H-s & ACC & AUC & H-s & ACC & AUC & H-s & ACC & AUC & H-s & ACC & AUC & H-score \\
\midrule
Source & 49.7 & 67.9 & 56.3 & 49.8 & 68.1 & 56.4 & 49.8 & 68.2 & 56.4 & 49.8 & 68.1 & 56.4 & 49.8 & 68.0 & 56.4 & 49.8 & 68.1 & 56.4 \\
Tent \cite{wang2021tent} & 51.2 & 65.3 & 55.0 & 49.2 & 60.3 & 51.8 & 30.0 & 54.4 & 31.9 & 29.1 & 53.8 & 31.3 & 22.4 & 50.9 & 23.8 & 36.4 & 56.9 & 38.8 \\
CoTTA \cite{wang2022continual} & 49.9 & 67.5 & 56.4 & 49.9 & 67.0 & 56.2 & 49.8 & 66.6 & 56.0 & 49.8 & 65.8 & 55.6 & 49.5 & 64.5 & 54.8 & 49.8 & 66.3 & 55.8 \\
EATA \cite{niu2022efficient} & 58.6 & 70.6 & 63.5 & \underbar{58.5} & 70.1 & 63.3 & 56.6 & 69.4 & 61.2 & 55.0 & 67.8 & 59.6 & 52.9 & 67.3 & 57.8 & 56.3 & 69.0 & 61.1 \\
SAR \cite{niu2023towards} & 57.1 & 71.9 & 63.2 & 56.1 & 69.2 & 61.4 & 54.0 & 66.8 & 58.7 & 52.8 & 64.5 & 57.2 & 50.4 & 61.5 & 54.3 & 54.1 & 66.8 & 59.0 \\
OSTTA \cite{lee2023towards} & 58.5 & 69.3 & 63.0 & 58.2 & 67.3 & 62.0 & \underbar{57.8} & 65.7 & 61.1 & \underbar{57.3} & 64.2 & 60.2 & 56.2 & 61.9 & 58.5 & \underbar{57.6} & 65.7 & 61.0 \\
ViDA \cite{liu2024vida} & 56.0 & 70.8 & 62.1 & 55.3 & 63.5 & 58.4 & 54.8 & 57.6 & 55.0 & 53.9 & 52.5 & 51.6 & 53.0 & 47.9 & 48.4 & 54.6 & 58.5 & 55.1 \\
UniEnt \cite{gao2024unified} & 57.6 & 75.1 & 64.5 & 57.7 & 75.5 & 64.7 & 57.8 & 76.2 & 65.1 & 56.8 & 76.3 & 64.2 & 57.8 & \underbar{77.0} & \underbar{65.4} & 57.5 & 76.0 & 64.8 \\
STAMP \cite{yu2024stamp} & 51.3 & 71.5 & 59.0 & 51.5 & 71.8 & 59.2 & 51.3 & 72.3 & 59.2 & 51.9 & 73.7 & 60.1 & 52.0 & 73.8 & 60.2 & 51.6 & 72.6 & 59.5 \\
E-COME \cite{zhang2025come} & 58.4 & 75.7 & 65.4 & 58.5 & 76.0 & 65.6 & 57.5 & 75.1 & 64.5 & 53.5 & 73.1 & 59.9 & \underbar{58.3} & 75.5 & 65.2 & 57.2 & 75.1 & 64.1 \\
S-COME \cite{zhang2025come} & 53.1 & 69.7 & 58.9 & 53.8 & 70.8 & 60.0 & 50.2 & 67.7 & 56.1 & 50.2 & 68.0 & 56.4 & 40.9 & 63.0 & 45.5 & 49.7 & 67.9 & 55.4 \\
DPCore \cite{zhang2025dpcore} & \underbar{59.0} & \underbar{79.3} & \underbar{67.2} & 58.3 & \underbar{80.1} & \underbar{66.9} & 56.8 & \underbar{78.5} & \underbar{65.4} & 56.2 & \underbar{78.3} & \underbar{64.9} & 54.1 & 76.2 & 62.6 & 56.9 & \underbar{78.5} & \underbar{65.4} \\
\rowcolor{blue!10}
\textbf{DOCO (Ours)} & \textbf{61.7} & \textbf{79.8} & \textbf{69.2} & \textbf{62.0} & \textbf{81.0} & \textbf{69.8} & \textbf{61.6} & \textbf{81.0} & \textbf{69.5} & \textbf{61.7} & \textbf{82.0} & \textbf{69.9} & \textbf{61.5} & \textbf{82.7} & \textbf{70.1} & \textbf{61.7} & \textbf{81.3} & \textbf{69.7} \\

\bottomrule
\end{tabular}
\label{tab:oodpercent_hs}
\end{table*}


\begin{table*}[t]
\centering
\small
\setlength{\tabcolsep}{3.8pt}
\caption{Results ($\%$) for ImageNet-to-ImageNet-C (severity = 5, $\kappa = 0.1$) in OCTTA setting across six covariate-shifted OOD datasets. }
\renewcommand{\arraystretch}{0.85}
\begin{tabular}{c cc cc cc cc cc cc ccc}
\toprule
\multirow{2}{*}{\textbf{Method}} &
\multicolumn{2}{c}{\textbf{Places.--C}} &
\multicolumn{2}{c}{\textbf{Texture--C}} &
\multicolumn{2}{c}{\textbf{iNatur.--C}} &
\multicolumn{2}{c}{\textbf{SUN--C}} &
\multicolumn{2}{c}{\textbf{SSB-H.--C}} &
\multicolumn{2}{c}{\textbf{NINCO--C}} &
\multicolumn{3}{c}{\textbf{Avg.}} \\
\cmidrule(lr){2-3} \cmidrule(lr){4-5} \cmidrule(lr){6-7} \cmidrule(lr){8-9} \cmidrule(lr){10-11} \cmidrule(lr){12-13} \cmidrule(lr){14-16}
& ACC & AUC & ACC & AUC & ACC & AUC & ACC & AUC & ACC & AUC & ACC & AUC & ACC & AUC & H-score \\
\midrule
Source & 49.7 & 66.5 & 49.7 & 70.7 & 49.7 & 78.4 & 49.7 & 71.6 & 49.7 & 56.1 & 49.7 & 64.4 & 49.7 & 67.9 & 56.3 \\
Tent \cite{wang2021tent} & 17.2 & 52.0 & 57.3 & 60.0 & 57.6 & 78.6 & 58.0 & 67.6 & 58.7 & \underbar{64.8} & 58.3 & 68.9 & 51.2 & 65.3 & 54.9 \\
CoTTA \cite{wang2022continual} & 49.9 & 65.7 & 49.9 & 69.9 & 49.9 & 77.9 & 49.9 & 70.7 & 50.0 & 56.4 & 50.0 & 64.2 & 49.9 & 67.5 & 56.4 \\
EATA \cite{niu2022efficient} & 56.6 & 66.3 & 59.5 & 74.7 & 58.2 & 78.6 & \underbar{59.3} & 73.8 & 58.4 & 61.8 & \underbar{59.8} & 68.2 & 58.6 & 70.6 & 63.5 \\
SAR \cite{niu2023towards} & 57.2 & 68.0 & 57.0 & 74.1 & 57.0 & 82.3 & 57.3 & 74.5 & 57.0 & 62.2 & 57.1 & 70.5 & 57.1 & 71.9 & 63.2 \\
OSTTA \cite{lee2023towards} & 58.4 & 65.7 & 58.3 & 66.1 & \underbar{58.5} & 79.7 & 58.5 & 71.8 & 58.7 & 63.7 & 58.7 & 68.7 & 58.5 & 69.3 & 63.0 \\
ViDA \cite{liu2024vida} & 56.1 & 71.6 & 55.5 & 69.2 & 55.9 & 74.2 & 55.7 & 69.1 & 56.4 & \textbf{66.6} & 56.0 & \underbar{73.9} & 55.9 & 70.8 & 62.1 \\
UniEnt \cite{gao2024unified} & 59.1 & 73.3 & 58.7 & 81.3 & 57.0 & 84.3 & 56.8 & 79.6 & 55.0 & 60.7 & 58.8 & 71.5 & 57.6 & 75.1 & 64.5 \\
STAMP \cite{yu2024stamp} & 51.2 & 71.0 & 51.1 & 71.3 & 51.4 & 82.9 & 51.4 & 74.8 & 51.5 & 59.7 & 51.2 & 69.2 & 51.3 & 71.5 & 59.0 \\
E-COME \cite{zhang2025come} & 58.4 & 75.9 & 58.2 & 78.4 & 58.4 & 85.3 & 58.4 & 82.1 & 58.2 & 60.5 & 58.7 & 72.0 & 58.4 & 75.7 & 65.4 \\
S-COME \cite{zhang2025come} & 42.4 & 65.8 & 54.3 & 74.4 & 54.0 & 75.7 & 55.3 & 76.6 & 55.4 & 58.5 & 57.1 & 67.4 & 53.1 & 69.7 & 58.9 \\
DPCore \cite{zhang2025dpcore} & \underbar{61.5} & \textbf{79.7} & \textbf{61.5} & \textbf{82.5} & 57.8 & \textbf{93.7} & 58.0 & \underbar{83.4} & \underbar{59.1} & 63.2 & 56.3 & 73.4 & \underbar{59.0} & \underbar{79.3} & \underbar{67.2} \\
\rowcolor{blue!10}
\textbf{DOCO (Ours)} & \textbf{61.7} & \underbar{76.8} & \underbar{61.1} & \underbar{82.5} & \textbf{61.9} & \underbar{92.3} & \textbf{62.4} & \textbf{88.0} & \textbf{62.0} & 64.4 & \textbf{61.3} & \textbf{74.8} & \textbf{61.7} & \textbf{79.8} & \textbf{69.2} \\
\bottomrule
\end{tabular}
\label{tab_sup:imgNetC_10}
\end{table*}

\begin{table*}[t]
\centering
\small
\setlength{\tabcolsep}{3.8pt}
\caption{Results ($\%$) for ImageNet-to-ImageNet-C (severity = 5, $\kappa = 0.2$) in OCTTA setting across six covariate-shifted OOD datasets.}
\renewcommand{\arraystretch}{0.85}
\begin{tabular}{c cc cc cc cc cc cc ccc}
\toprule
\multirow{2}{*}{\textbf{Method}} &
\multicolumn{2}{c}{\textbf{Places.--C}} &
\multicolumn{2}{c}{\textbf{Texture--C}} &
\multicolumn{2}{c}{\textbf{iNatur.--C}} &
\multicolumn{2}{c}{\textbf{SUN--C}} &
\multicolumn{2}{c}{\textbf{SSB-H.--C}} &
\multicolumn{2}{c}{\textbf{NINCO--C}} &
\multicolumn{3}{c}{\textbf{Avg.}} \\
\cmidrule(lr){2-3} \cmidrule(lr){4-5} \cmidrule(lr){6-7} \cmidrule(lr){8-9} \cmidrule(lr){10-11} \cmidrule(lr){12-13} \cmidrule(lr){14-16}
& ACC & AUC & ACC & AUC & ACC & AUC & ACC & AUC & ACC & AUC & ACC & AUC & ACC & AUC & H-score \\
\midrule
Source & 49.8 & 67.0 & 49.8 & 70.8 & 49.8 & 78.3 & 49.8 & 71.6 & 49.8 & 56.2 & 49.8 & 64.5 & 49.8 & 68.1 & 56.4 \\
Tent \cite{wang2021tent} & 56.9 & 61.8 & 31.2 & 50.7 & 33.8 & 57.5 & 56.8 & 58.0 & 58.5 & \underbar{65.5} & 57.9 & 68.3 & 49.2 & 60.3 & 51.8 \\
CoTTA \cite{wang2022continual} & 49.9 & 65.6 & 49.9 & 69.0 & 49.7 & 77.3 & 50.0 & 69.8 & 49.9 & 56.3 & 49.9 & 64.0 & 49.9 & 67.0 & 56.2 \\
EATA \cite{niu2022efficient} & 57.9 & 66.5 & 58.5 & 73.1 & 58.7 & 78.4 & 58.8 & 73.4 & 59.2 & 62.3 & 57.7 & 66.9 & 58.5 & 70.1 & 63.3 \\
SAR \cite{niu2023towards} & 56.9 & 67.5 & 53.3 & 67.8 & 56.0 & 78.9 & 56.3 & 68.1 & 57.3 & 62.9 & 56.8 & 70.0 & 56.1 & 69.2 & 61.4 \\
OSTTA \cite{lee2023towards} & 58.0 & 64.5 & 57.7 & 62.3 & 58.0 & 76.2 & 58.4 & 67.8 & 58.5 & 64.6 & 58.3 & 68.3 & 58.2 & 67.3 & 62.0 \\
ViDA \cite{liu2024vida} & 55.7 & 69.5 & 54.5 & 56.7 & 55.1 & 59.6 & 55.0 & 57.4 & 56.2 & \textbf{66.3} & 55.5 & 71.6 & 55.3 & 63.5 & 58.4 \\
UniEnt \cite{gao2024unified} & 59.1 & 73.9 & \underbar{59.5} & 82.1 & 52.2 & 82.6 & \underbar{59.4} & 82.7 & 56.7 & 59.8 & \underbar{59.2} & 71.8 & 57.7 & 75.5 & 64.7 \\
STAMP \cite{yu2024stamp} & 51.6 & 71.6 & 51.3 & 71.6 & 51.4 & 83.0 & 51.4 & 74.7 & 51.6 & 60.0 & 51.5 & 69.6 & 51.5 & 71.7 & 59.2 \\
E-COME \cite{zhang2025come} & \underbar{59.6} & 76.7 & 59.1 & 79.1 & \underbar{59.1} & 86.4 & 56.9 & 81.8 & 58.7 & 60.2 & 57.9 & 72.1 & \underbar{58.6} & 76.1 & 65.6 \\
S-COME \cite{zhang2025come} & 54.6 & 72.1 & 44.9 & 73.1 & 55.5 & 77.2 & 56.8 & 77.2 & 55.8 & 58.4 & 55.5 & 66.9 & 53.8 & 70.8 & 60.0 \\
DPCore \cite{zhang2025dpcore} & 58.4 & \textbf{79.7} & 58.4 & \underbar{82.8} & 57.0 & \underbar{93.3} & 58.4 & \underbar{87.6} & \underbar{60.4} & 63.1 & 57.1 & \underbar{74.4} & 58.3 & \underbar{80.1} & \underbar{66.9} \\
\rowcolor{blue!10}
\textbf{DOCO (Ours)} & \textbf{62.1} & \underbar{78.4} & \textbf{61.8} & \textbf{83.6} & \textbf{62.5} & \textbf{94.0} & \textbf{61.9} & \textbf{90.0} & \textbf{61.9} & 64.7 & \textbf{62.1} & \textbf{75.2} & \textbf{62.0} & \textbf{81.0} & \textbf{69.8} \\

\bottomrule
\end{tabular}
\label{tab_sup:imgNetC_20}
\end{table*}

\begin{table*}[t]
\centering
\small
\setlength{\tabcolsep}{3.8pt}
\caption{Results ($\%$) for ImageNet-to-ImageNet-C (severity = 5, $\kappa = 0.3$) in OCTTA setting across six covariate-shifted OOD datasets. }
\renewcommand{\arraystretch}{0.85}
\begin{tabular}{c cc cc cc cc cc cc ccc}
\toprule
\multirow{2}{*}{\textbf{Method}} &
\multicolumn{2}{c}{\textbf{Places.--C}} &
\multicolumn{2}{c}{\textbf{Texture--C}} &
\multicolumn{2}{c}{\textbf{iNatur.--C}} &
\multicolumn{2}{c}{\textbf{SUN--C}} &
\multicolumn{2}{c}{\textbf{SSB-H.--C}} &
\multicolumn{2}{c}{\textbf{NINCO--C}} &
\multicolumn{3}{c}{\textbf{Avg.}} \\
\cmidrule(lr){2-3} \cmidrule(lr){4-5} \cmidrule(lr){6-7} \cmidrule(lr){8-9} \cmidrule(lr){10-11} \cmidrule(lr){12-13} \cmidrule(lr){14-16}
& ACC & AUC & ACC & AUC & ACC & AUC & ACC & AUC & ACC & AUC & ACC & AUC & ACC & AUC & H-score \\
\midrule
Source & 49.8 & 67.0 & 49.8 & 71.1 & 49.8 & 78.7 & 49.8 & 71.7 & 49.8 & 56.0 & 49.8 & 64.5 & 49.8 & 68.2 & 56.4 \\
Tent \cite{wang2021tent} & 18.2 & 53.5 & 11.3 & 46.4 & 34.3 & 58.1 & 41.6 & 46.6 & 17.7 & 54.4 & 56.9 & 67.6 & 30.0 & 54.4 & 31.9 \\
CoTTA \cite{wang2022continual} & 49.8 & 65.2 & 49.7 & 68.3 & 49.8 & 77.5 & 50.0 & 69.1 & 50.1 & 55.9 & 49.6 & 63.5 & 49.8 & 66.6 & 55.9 \\
EATA \cite{niu2022efficient} & 56.0 & 66.2 & 56.6 & 72.2 & 55.4 & 76.7 & 54.8 & 72.8 & 58.1 & 60.8 & \underbar{58.6} & 67.4 & 56.6 & 69.4 & 61.2 \\
SAR \cite{niu2023towards} & 54.8 & 66.1 & 50.0 & 63.6 & 54.2 & 78.7 & 52.8 & 61.5 & 57.3 & 62.2 & 54.7 & 68.9 & 54.0 & 66.8 & 58.7 \\
OSTTA \cite{lee2023towards} & 58.1 & 63.2 & 57.1 & 59.8 & 56.6 & 74.3 & 57.7 & 65.0 & \underbar{58.8} & 63.9 & 58.4 & 68.0 & \underbar{57.8} & 65.7 & 61.1 \\
ViDA \cite{liu2024vida} & 55.1 & 66.7 & 54.0 & 48.7 & 54.6 & 50.0 & 53.9 & 46.7 & 56.1 & \textbf{64.5} & 55.1 & 69.2 & 54.8 & 57.6 & 55.0 \\
UniEnt \cite{gao2024unified} & 56.1 & 73.2 & \underbar{59.0} & \underbar{82.8} & \underbar{59.8} & \underbar{89.8} & 57.1 & 81.2 & 56.6 & 58.7 & 58.1 & 71.8 & 57.8 & 76.2 & 65.1 \\
STAMP \cite{yu2024stamp} & 51.1 & 71.8 & 51.0 & 72.7 & 51.5 & 83.9 & 51.2 & 75.6 & 51.7 & 60.0 & 51.4 & 70.0 & 51.3 & 72.3 & 59.2 \\
E-COME \cite{zhang2025come} & \underbar{58.9} & \underbar{76.5} & 57.3 & 78.1 & 54.5 & 82.2 & \underbar{58.0} & 82.6 & 58.1 & 59.0 & 58.4 & 72.1 & 57.5 & 75.1 & 64.5 \\
S-COME \cite{zhang2025come} & 29.9 & 57.1 & 55.4 & 75.5 & 54.2 & 75.1 & 55.0 & 76.3 & 53.4 & 56.4 & 53.7 & 66.1 & 50.2 & 67.7 & 56.1 \\
DPCore \cite{zhang2025dpcore} & 57.5 & 76.1 & 55.1 & 80.5 & 57.0 & 89.2 & 57.3 & \underbar{86.4} & 58.0 & 63.5 & 55.8 & \underbar{75.1} & 56.8 & \underbar{78.5} & \underbar{65.4} \\
\rowcolor{blue!10}
\textbf{DOCO (Ours)} & \textbf{61.4} & \textbf{79.0} & \textbf{61.6} & \textbf{83.4} & \textbf{61.6} & \textbf{94.3} & \textbf{61.5} & \textbf{89.5} & \textbf{62.1} & \underbar{64.4} & \textbf{61.4} & \textbf{75.3} & \textbf{61.6} & \textbf{81.0} & \textbf{69.5} \\

\bottomrule
\end{tabular}
\label{tab_sup:imgNetC_30}
\end{table*}

\begin{table*}[t]
\centering
\small
\setlength{\tabcolsep}{3.8pt}
\caption{Results ($\%$) for ImageNet-to-ImageNet-C (severity = 5, $\kappa = 0.4$) in OCTTA setting across six covariate-shifted OOD datasets. }
\renewcommand{\arraystretch}{0.85}
\begin{tabular}{c cc cc cc cc cc cc ccc}
\toprule
\multirow{2}{*}{\textbf{Method}} &
\multicolumn{2}{c}{\textbf{Places.--C}} &
\multicolumn{2}{c}{\textbf{Texture--C}} &
\multicolumn{2}{c}{\textbf{iNatur.--C}} &
\multicolumn{2}{c}{\textbf{SUN--C}} &
\multicolumn{2}{c}{\textbf{SSB-H.--C}} &
\multicolumn{2}{c}{\textbf{NINCO--C}} &
\multicolumn{3}{c}{\textbf{Avg.}} \\
\cmidrule(lr){2-3} \cmidrule(lr){4-5} \cmidrule(lr){6-7} \cmidrule(lr){8-9} \cmidrule(lr){10-11} \cmidrule(lr){12-13} \cmidrule(lr){14-16}
& ACC & AUC & ACC & AUC & ACC & AUC & ACC & AUC & ACC & AUC & ACC & AUC & ACC & AUC & H-score \\
\midrule
Source & 49.8 & 66.9 & 49.8 & 70.9 & 49.8 & 78.6 & 49.8 & 71.7 & 49.8 & 56.2 & 49.8 & 64.3 & 49.8 & 68.1 & 56.4 \\
Tent \cite{wang2021tent} & 26.1 & 51.8 & 6.0 & 53.1 & 14.3 & 47.2 & 17.2 & 41.7 & 58.2 & 64.3 & 52.4 & 64.8 & 29.0 & 53.8 & 31.3 \\
CoTTA \cite{wang2022continual} & 49.8 & 64.2 & 49.4 & 67.3 & 49.8 & 76.6 & 49.9 & 67.8 & 50.1 & 55.9 & 49.6 & 62.8 & 49.8 & 65.8 & 55.6 \\
EATA \cite{niu2022efficient} & 55.8 & 65.6 & 52.8 & 69.5 & 54.8 & 75.5 & 54.6 & 72.1 & 56.3 & 58.1 & 55.8 & 66.0 & 55.0 & 67.8 & 59.6 \\
SAR \cite{niu2023towards} & 54.6 & 64.8 & 44.8 & 59.0 & 53.5 & 70.3 & 52.8 & 63.2 & 57.0 & 61.8 & 54.4 & 68.1 & 52.8 & 64.5 & 57.2 \\
OSTTA \cite{lee2023towards} & \underbar{58.0} & 62.1 & 56.7 & 57.7 & 54.9 & 71.3 & \underbar{57.1} & 62.6 & 58.8 & 63.8 & \underbar{58.2} & 67.5 & \underbar{57.3} & 64.2 & 60.2 \\
ViDA \cite{liu2024vida} & 54.0 & 63.2 & 52.9 & 40.4 & 53.9 & 42.5 & 52.7 & 38.7 & 55.8 & 63.6 & 54.0 & 66.5 & 53.9 & 52.5 & 51.6 \\
UniEnt \cite{gao2024unified} & 57.6 & 72.7 & 56.6 & 81.1 & 56.3 & \underbar{88.2} & 55.9 & 82.7 & 58.9 & 62.2 & 55.3 & 71.0 & 56.7 & 76.3 & 64.1 \\
STAMP \cite{yu2024stamp} & 52.0 & 72.5 & 51.9 & 75.0 & 51.8 & 85.4 & 51.9 & 77.3 & 51.9 & 60.7 & 52.0 & 71.1 & 51.9 & 73.7 & 60.1 \\
E-COME \cite{zhang2025come} & 54.8 & 74.6 & 56.9 & 78.0 & \underbar{59.2} & 86.2 & 34.4 & 70.3 & 59.1 & 59.1 & 56.4 & 70.7 & 53.5 & 73.1 & 59.9 \\
S-COME \cite{zhang2025come} & 54.9 & 71.6 & 53.1 & 73.5 & 53.9 & 76.2 & 36.7 & 64.5 & 49.7 & 56.3 & 53.2 & 65.7 & 50.2 & 68.0 & 56.4 \\
DPCore \cite{zhang2025dpcore} & 56.9 & \underbar{77.3} & \underbar{58.7} & \underbar{82.5} & 50.6 & 86.5 & 54.9 & \underbar{83.5} & \underbar{60.0} & \underbar{64.6} & 56.5 & \underbar{75.6} & 56.2 & \underbar{78.3} & \underbar{64.9} \\
\rowcolor{blue!10}
\textbf{DOCO (Ours)} & \textbf{62.4} & \textbf{80.1} & \textbf{62.1} & \textbf{84.6} & \textbf{61.3} & \textbf{95.5} & \textbf{60.5} & \textbf{90.8} & \textbf{61.9} & \textbf{65.2} & \textbf{62.1} & \textbf{75.8} & \textbf{61.7} & \textbf{82.0} & \textbf{69.9} \\
\bottomrule
\end{tabular}
\label{tab_sup:imgNetC_40}
\end{table*}

\begin{table*}[t]
\centering
\small
\setlength{\tabcolsep}{3.8pt}
\caption{Results ($\%$) for LAION-C benchmark (severity = 1, $\kappa = 0.5$) in OCTTA setting across six covariate-shifted OOD datasets. All the results are averaged over 6 constantly switching domains. \texttt{-L} stands for applying LAION-C corruption to OOD dataset. }
\renewcommand{\arraystretch}{0.85}
\begin{tabular}{c cc cc cc cc cc cc ccc}
\toprule
\multirow{2}{*}{\textbf{Method}} &
\multicolumn{2}{c}{\textbf{Places.--L}} &
\multicolumn{2}{c}{\textbf{Texture--L}} &
\multicolumn{2}{c}{\textbf{iNatur.--L}} &
\multicolumn{2}{c}{\textbf{SUN--L}} &
\multicolumn{2}{c}{\textbf{SSB-H.--L}} &
\multicolumn{2}{c}{\textbf{NINCO--L}} &
\multicolumn{3}{c}{\textbf{Avg.}} \\
\cmidrule(lr){2-3} \cmidrule(lr){4-5} \cmidrule(lr){6-7} \cmidrule(lr){8-9} \cmidrule(lr){10-11} \cmidrule(lr){12-13} \cmidrule(lr){14-16}
& ACC & AUC & ACC & AUC & ACC & AUC & ACC & AUC & ACC & AUC & ACC & AUC & ACC & AUC & H-score \\
\midrule
Source & 51.4 & 63.8 & 51.4 & 68.9 & 51.4 & 77.7 & 51.4 & 72.6 & 51.4 & 60.8 & 51.4 & 66.5 & 51.4 & 68.4 & 57.2 \\
Tent (ICLR'21) & 2.5 & 47.9 & 2.8 & 50.9 & 1.6 & 39.6 & 1.7 & 54.1 & 2.5 & 53.8 & 2.1 & 51.9 & 2.2 & 49.7 & 3.5 \\
CoTTA (CVPR'22) & 51.3 & 61.6 & 51.3 & 65.0 & 51.3 & 72.4 & 51.2 & 68.8 & 51.3 & 60.8 & 50.4 & 64.1 & 51.1 & 65.4 & 55.9 \\
EATA (ICML'22) & 61.7 & 64.5 & 61.6 & 72.4 & 59.9 & 78.5 & 60.7 & 77.0 & 61.9 & 62.6 & 61.9 & 65.2 & 61.3 & 70.0 & 64.9 \\
SAR (ICLR'23) & 29.1 & 52.2 & 39.6 & 56.3 & 31.6 & 57.5 & 32.4 & 58.0 & 52.2 & 61.8 & 51.5 & 66.7 & 39.4 & 58.7 & 44.4 \\
OSTTA (ICCV'23) & 56.8 & 58.1 & 54.6 & 55.1 & 52.2 & 60.5 & 54.5 & 60.2 & 57.9 & 63.1 & 57.2 & 64.1 & 55.5 & 60.2 & 57.0 \\
ViDA (ICLR'24) & 30.8 & 46.8 & 33.6 & 45.0 & 29.1 & 37.4 & 29.5 & 42.1 & 35.0 & 62.2 & 29.7 & 55.3 & 31.3 & 48.1 & 34.2 \\
UniEnt (CVPR'24) & 61.4 & 70.6 & 61.2 & \underbar{82.1} & \underbar{61.0} & \underbar{88.2} & 61.2 & 85.0 & 60.7 & 64.0 & 60.8 & 69.3 & 61.0 & 76.5 & 67.3 \\
STAMP (ECCV'24) & 51.3 & 68.0 & 51.2 & 67.7 & 51.2 & 75.1 & 51.3 & 70.9 & 51.2 & 62.2 & 51.2 & 67.8 & 51.2 & 68.6 & 57.7 \\
E-COME (ICLR'25) & 60.5 & 73.3 & 48.7 & 72.2 & 59.9 & 86.5 & 61.0 & 81.4 & 60.9 & 59.6 & 59.9 & 70.1 & 58.5 & 73.8 & 64.2 \\
S-COME (ICLR'25) & 1.8 & 46.3 & 12.4 & 53.9 & 16.2 & 53.2 & 1.5 & 55.1 & 59.6 & 57.0 & 48.1 & 61.9 & 23.3 & 54.6 & 25.1 \\
DPCore (ICML'25) & \underbar{65.2} & \textbf{78.8} & \underbar{62.3} & 80.4 & 52.9 & 86.5 & \underbar{62.9} & \underbar{87.1} & \underbar{64.6} & \underbar{67.8} & \underbar{62.3} & \textbf{75.9} & \underbar{61.7} & \underbar{79.4} & \underbar{68.9} \\
\rowcolor{blue!10}
\textbf{DOCO (Ours)} & \textbf{66.9} & \underbar{77.6} & \textbf{65.7} & \textbf{82.4} & \textbf{64.9} & \textbf{94.9} & \textbf{65.4} & \textbf{89.7} & \textbf{66.2} & \textbf{68.8} & \textbf{64.7} & \underbar{75.7} & \textbf{65.6} & \textbf{81.5} & \textbf{72.3} \\
\bottomrule
\end{tabular}
\label{tab:sup_imgNetL_sev1}
\end{table*}

\subsection{Different severity experiment}
\label{sec:sup_laionsev1}
Similarly, we test our method on the LAION-C benchmark with a lower corruption severity level of 1, while keeping the OOD ratio at $\kappa=0.5$. As shown in \cref{tab:sup_imgNetL_sev1}, DOCO continues to outperform other methods, securing the highest average metrics, surpassing the second by $\bm{3.4\%}$. This demonstrates that DOCO's effectiveness is not limited to extreme domain shifts but also holds in scenarios with more subtle corruptions, confirming its consistent superiority.

\subsection{Different OOD Score Measurement}
\label{sec:sup_other_oodscore}

\begin{table}[h]
    \centering
    \setlength{\tabcolsep}{4pt} 
    \begin{tabular}{cccccc}
        \toprule
        \multirow{2}{*}{Method} & \multicolumn{4}{c}{OOD score} & \multirow{2}{*}{Mean$\pm$Std} \\
        \cmidrule(lr){2-5}
        & Ent & MLS & Energy & MSP & \\
        \midrule
Source                        & 57.29 & 57.02 & 56.37 & 56.96 & 56.91$\pm$0.39 \\
Tent \cite{wang2021tent}      & 24.25 & 23.99 & 23.82 & 24.11 & 24.04$\pm$0.18 \\
CoTTA \cite{wang2022continual}& 56.25 & 55.42 & 54.79 & 55.82 & 55.57$\pm$0.62 \\
EATA \cite{niu2022efficient}  & 57.98 & 57.84 & 57.77 & 57.29 & 57.72$\pm$0.30 \\
SAR \cite{niu2023towards}     & 55.10 & 54.51 & 54.31 & 54.63 & 54.64$\pm$0.34 \\
OSTTA \cite{lee2023towards}   & 60.39 & 58.92 & 58.51 & 59.82 & 59.41$\pm$0.85 \\
ViDA \cite{liu2024vida}       & 48.03 & 48.22 & 48.40 & 47.92 & 48.14$\pm$0.21 \\
UniEnt \cite{gao2024unified}  & 64.95 & \underline{65.25} & \underline{65.39} & 64.02 & 64.90$\pm$0.62 \\
STAMP \cite{yu2024stamp}      & 60.16 & 59.94 & 60.18 & 59.94 & 60.05$\pm$0.13 \\
E-COME \cite{zhang2025come}   & \underline{65.36} & 65.00 & 65.22 & \underline{64.73} & \underline{65.08$\pm$0.27} \\
S-COME \cite{zhang2025come}   & 45.87 & 45.53 & 45.47 & 45.37 & 45.56$\pm$0.22 \\
DPCore \cite{zhang2025dpcore} & 62.12 & 61.76 & 62.62 & 61.05 & 61.89$\pm$0.66 \\
\rowcolor{blue!10}
\textbf{DOCO (Ours)}                       & \textbf{69.57} & \textbf{69.38} & \textbf{70.10} & \textbf{68.45} & \textbf{69.38$\pm$0.69} \\
        \bottomrule
    \end{tabular}
    \caption{H-score results on ImageNet-C with $\kappa=0.5,sev =5$.}
    \label{tab:imagenet_ood_hscore_5000}
\end{table}

In the main paper, we adopt the energy-based OOD score as the default choice for computing AUC and H-score. To verify that our conclusions are not tied to a particular score, we further evaluate all methods under three additional mainstream post-hoc OOD scores, including entropy, Max Logit (MLS), and maximum softmax probability (MSP). As summarized in \cref{tab:imagenet_ood_hscore_5000}, DOCO consistently achieves the best H-score under all four score functions and exhibits only minor variation across them, whereas the strongest competing method reaches at most $65.36\%$. These results indicate that DOCO is insensitive to the specific OOD score used for evaluation and remains clearly ahead of existing baselines across different OOD score measurements.


\end{document}